\documentclass{article}

 \usepackage[preprint]{neurips_2026}

\usepackage{graphicx}
\usepackage{caption}
\usepackage{subcaption}
\usepackage{array}
\usepackage{booktabs}
\usepackage[table]{xcolor}
\usepackage{placeins}
\usepackage{amsmath}
\usepackage{amsfonts}
\usepackage{amssymb}
\usepackage{multirow}

\usepackage{ulem}
\usepackage[utf8]{inputenc}
\usepackage[T1]{fontenc}
\usepackage{hyperref}
\usepackage{url}
\usepackage{nicefrac}
\usepackage{microtype}
\usepackage{wrapfig}
\usepackage{algorithm}
\usepackage{algpseudocode}

\usepackage{amssymb}
\usepackage{pifont}
\newcommand{\cmark}{{\color{green}\ding{51}}}
\newcommand{\xmark}{{\color{red}\ding{55}}}

\newcommand{\zhd}[1]{{\color{black} {#1}}}
\newcommand{\xu}[1]{{\color{black} {#1}}}

\title{Kalman Meets Curriculum: Efficient Dynamic Prompt Selection for Adaptive RL Finetuning}

\author{
  \textbf{Haodong Zhu}\textsuperscript{1,2}\thanks{Equal contribution.} \quad
  \textbf{Yangyang Ren}\textsuperscript{1,2}\footnotemark[1] \quad
  \textbf{Yanjing Li}\textsuperscript{3}\thanks{Corresponding author.} \quad
  \textbf{Sheng Xu}\textsuperscript{4}\footnotemark[2] \\
  \textbf{Haiguang Liu}\textsuperscript{2} \quad
  \textbf{Linlin Yang}\textsuperscript{3} \quad
  \textbf{Baochang Zhang}\textsuperscript{1} \thanks{project lead.} \\[6pt]
  \textsuperscript{1}Beihang University \quad
  \textsuperscript{2}Zhongguancun Academy \quad
  \textsuperscript{3}Nanyang Technological university \\
  \textsuperscript{4}Independent Researcher \quad
  \textsuperscript{5}Communication University of China \\[2pt]
  \texttt{HaodongZhu@buaa.edu.cn} \quad
  \texttt{yyren@buaa.edu.cn}
}

\begin{document}

\maketitle

\begin{abstract}
Reinforcement learning (RL) finetuning significantly enhances the reasoning capabilities of large language models (LLMs), yet \xu{its effectiveness critically depends on selecting prompts of appropriate difficulty  for the current policy.}
\xu{This is challenging because prompt difficulty evolves throughout training.}
\xu{Existing online methods therefore face a trade-off: evaluation-based approaches are accurate but expensive, while prediction-based approaches are efficient but typically assume stationary difficulty, making them ill-suited to RL’s non-stationary training dynamics.}
To address these issues, we propose a \emph{\textbf{K}alman-\textbf{G}uided \textbf{P}rompt \textbf{S}election} method (\emph{\textbf{KGPS}}), which reformulates prompt selection as a dynamic state estimation problem \xu{rather than static difficulty prediction.}
\xu{KGPS models each prompt’s latent success rate in logit space using a linear-Gaussian state-space model, with process noise coupled to the magnitude of policy updates so that uncertainty increases when the policy changes more substantially.}
\xu{A Kalman filter then maintains a calibrated Gaussian posterior over prompt difficulty, and prompts are selected by maximizing a posterior-expected training utility that favors intermediate-difficulty prompts while naturally revisiting uncertain ones.}
\xu{The resulting procedure is adaptive to policy drift and requires no additional rollouts beyond standard policy training.}
\xu{Extensive experiments across mathematics, planning, and geometry reasoning benchmarks, as well as multiple RL algorithms, show that KGPS consistently improves both final accuracy and rollout efficiency over strong baselines, establishing state-of-the-art performance among online prompt selection methods. For example, on DeepSeek-R1-Distill-7B, KGPS uses 83\% fewer rollouts than DS while even improving the average performance by 0.12 point across six math reasoning benchmarks.}
\end{abstract}

\begin{figure}[ht]
    \centering
    \begin{minipage}{0.48\linewidth}
        \centering
        \includegraphics[width=\linewidth, height=0.22\textheight]{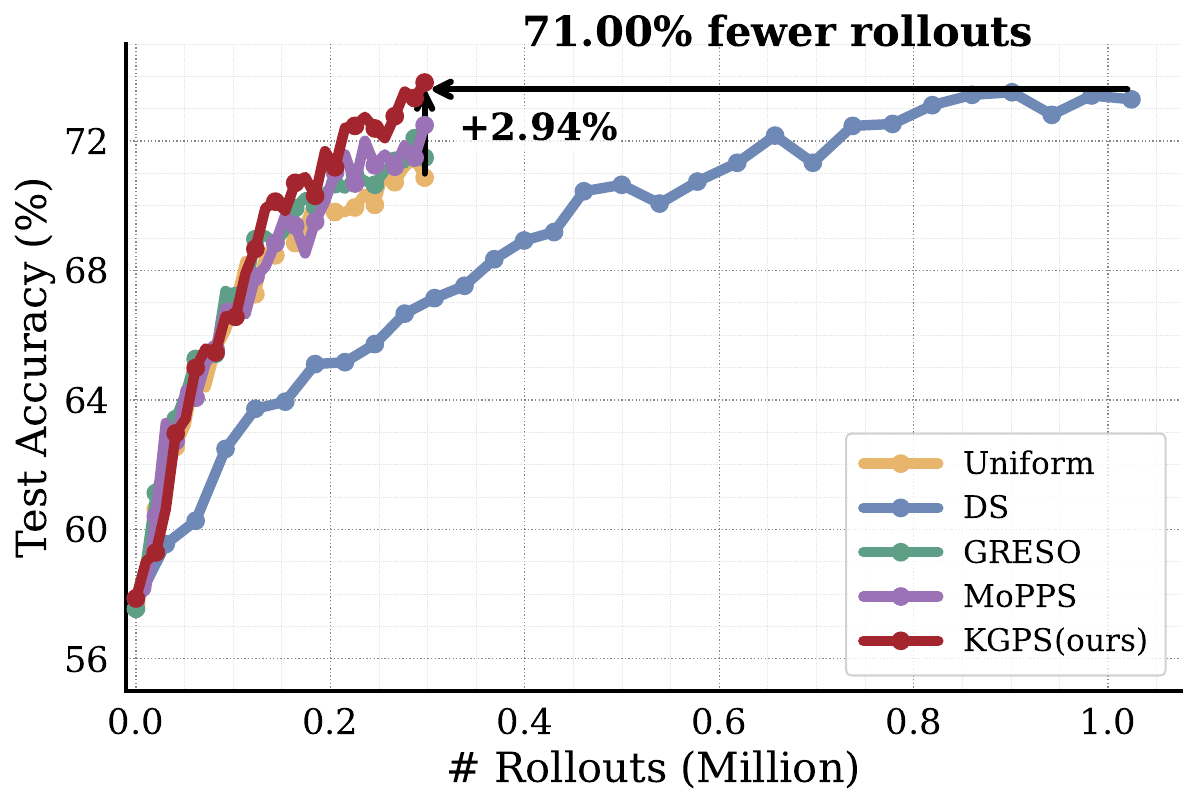}
        {\small (a) Test Accuracy vs.\ Rollout Cost%把子标题写了，简短描述，和图1总标题有对应
        }
    \end{minipage}
    \hfill
    \begin{minipage}{0.48\linewidth}
        \centering
        \vspace{10pt}
        \includegraphics[width=\linewidth, height=0.22\textheight]{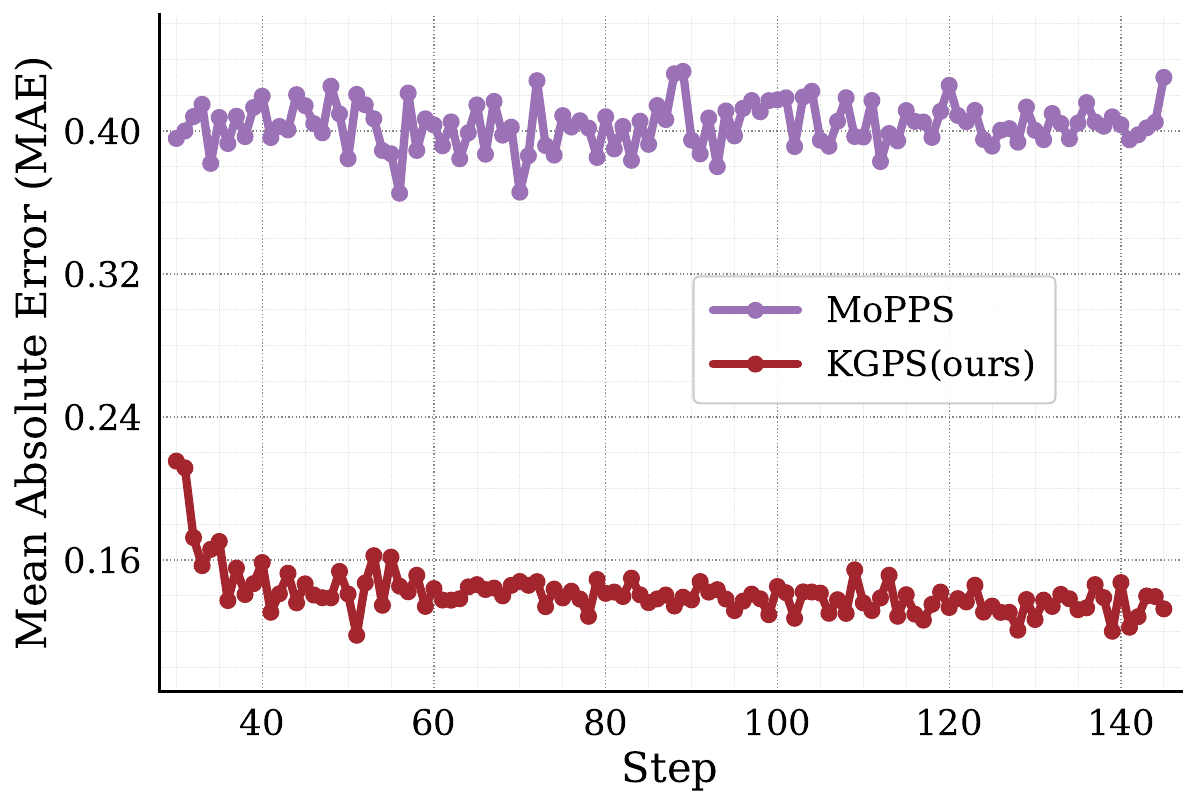}
        {\small (b) \xu{Evolution of MAE between empirical success-rate and predicted success-rates over training}%把子标题写了，简短描述，和图1总标题有对应
        }
    \end{minipage}
    \caption{On the Math dataset using the Qwen3-0.6B model, (a) \xu{KGPS obtains comparable performance to evaluation-based DS (Oracle) with 71.00\% fewer rollouts, also surpasses vanilla uniform method by up to +2.94\% higher test accuracy with the same rollout cost. Meanwhile, compared to prediction-based prompt selection methods GRESO and MoPPS, 
    the proposed KGPS achieves up to +2.33\% higher test accuracy.}
    (b) \xu{We calculate the MAE between empirical success-rate and predicted success-rate, where}
    KGPS maintains substantially lower \xu{MAE} than MoPPS throughout training, demonstrating superior prompt difficulty prediction without additional LLM inference.}
    \label{fig:head}
    \vspace{-19pt}
\end{figure}

\section{Introduction}

Reinforcement learning (RL) has become a cornerstone technique for post-training Large Language Models (LLMs), driving remarkable advances in both instruction following and complex reasoning \citep{shao2024deepseekmath, deepseekai2025deepseekr1incentivizingreasoningcapability, zeng2025simplerlzooinvestigatingtamingzero, deepscaler2025}.
However, realizing its full potential remains non-trivial. 
RL optimization is known to be highly sensitive to training sample selection \citep{parashar2025curriculumreinforcementlearningeasy, qu2025promptdifficultyonlinepredicted, shen2025bots, shen2025skyworkr1v3technicalreport}, \xu{because prompts that are already mastered by the current policy contribute little learning signal, while prompts that remain far beyond the policy’s capability often produce uninformative gradients.}
%
% and policy evaluation in GRPO-style frameworks \citep{deepseekai2025deepseekr1incentivizingreasoningcapability} requires massive rollout generation, introducing considerable computational and memory overhead. 
%
% These challenges are further compounded by static, uniformly sampled training distributions, where rollouts expended on already-mastered or currently-intractable problems contribute negligible learning signal, reduce effective batch diversity, and ultimately destabilize optimization 
\xu{In practice, the most useful prompts tend to be those at an appropriate difficulty for the current policy. Crucially, prompt difficulty is not fixed: as the policy improves, previously informative prompts may become trivial, and the set of useful training samples correspondingly shifts over time. This dynamic undermines static selection strategies—including uniform sampling and fixed curricula—which can waste rollout budget on uninformative prompts and become increasingly misaligned with the model’s evolving competence~\citep{yu2025dapo, bae2025online}.}
\xu{This underscores the need for adaptive task selection that remains aligned with the current policy throughout training.}
% the necessity of adaptive task selection, which dynamically curates training samples at appropriate difficulty levels to sustain informative gradients as the model's capabilities evolve.
% Such misalignment

% Prior work on data selection for RL training can be broadly categorized into offline and online approaches. 
% Offline methods precompute static difficulty estimates prior to training and organize samples into a fixed curriculum progressing from easy to hard \citep{parashar2025curriculumreinforcementlearningeasy, wen2025saristructuredaudioreasoning, shen2025skyworkr1v3technicalreport}. 
% While effective in the early stages of training, such static orderings become increasingly misaligned with the model's actual capability as the policy improves, fundamentally limiting their ability to sustain efficient learning throughout the full training process. 
% This limitation has motivated the development of online data selection methods \citep{qu2025promptdifficultyonlinepredicted, shen2025bots, yu2025dapoopensourcellmreinforcement}, which dynamically recompute sample utility at each update step using real-time feedback from the current policy, thereby maintaining a tighter alignment between the selected data and the model's evolving competence. \xu{In RL finetuning, this online setting is especially important because prompt success rate under the current policy is itself a non-stationary quantity.}

Prior work on data selection for RL training splits into offline and online approaches. Offline methods precompute static difficulty estimates and use fixed curricula~\citep{parashar2025curriculumreinforcementlearningeasy, wen2025saristructuredaudioreasoning, shen2025skyworkr1v3technicalreport}, but these misalign as the policy improves. Online methods dynamically recompute sample utility using real-time feedback~\citep{qu2025promptdifficultyonlinepredicted, shen2025bots, yu2025dapo}, maintaining alignment with evolving competence. \xu{In RL finetuning, this online setting is especially important because prompt success rate under the current policy is itself non-stationary.}

Online selection methods broadly fall into two categories depending on how sample utility is estimated. 
Evaluation-based approaches, such as Dynamic Sampling (DS)~\citep{yu2025dapo}, operate \xu{by oversampling a large candidate set, executing rollouts across all candidates, and filtering prompts based on the observed reward outcomes.}
% after rollout generation. At each training step, they oversample a large candidate set, execute rollouts across all candidates, and filter based on the observed reward outcomes. 
%
While yielding reliable utility estimates, \xu{these candidate evaluations} demand \xu{substantial additional} rollouts \xu{beyond those already needed for policy training}, resulting in significant inference overhead that compounds at every update step. 
% 这里点明是哪个方法、哪几条曲线对比得出的结论，我读到这里都没看到DS对应哪篇文章
As illustrated in Fig.~\ref{fig:head} (a), \xu{in DS training}, this overhead \xu{(blue curve)} can reach up to $7\times$ more rollouts than standard GRPO training \xu{(yellow curve)}, severely limiting practical scalability.
To address this, prediction-based approaches, notably MoPPS~\citep{qu2025promptdifficultyonlinepredicted} \xu{(purple curve)} and GRESO~\citep{greso} \xu{(green curve)}, rely on lightweight proxy signals to identify prompts near a target difficulty level without requiring additional rollouts \xu{for selection}. 
However, they model each prompt's success rate as a stationary latent parameter, relying on a heuristic decay factor as a substitute for principled non-stationarity handling, 
\zhd{which may yield inaccurate difficulty estimates as the policy evolves.}
%}
\zhd{To quantify this estimation inaccuracy, we calculate the MAE between predicted success-rate (before rollouts) and empirical success-rate (after rollouts) at every training step in Fig.~\ref{fig:head} (b).}
% \xu{To evaluate such representation gap, we calculate the MAE between predicted success-rate (before rollouts) and empirical success-rate (after rollouts) at every training step in Fig.~\ref{fig:head} (b).}
%
\xu{As can be seen}, this static modeling~\cite{qu2025promptdifficultyonlinepredicted} assumption results in persistently high estimation error throughout training ($\text{MAE} \approx 0.40$ \xu{in purple curve}), failing to improve as the policy evolves.
\textbf{These limitations call for a principled shift from static parameter estimation to dynamic state tracking}, \xu{one that preserves the efficiency of inference-free selection while better matching the non-stationary nature of RL finetuning.}
% one that coherently propagates uncertainty over time and derives selection criteria directly from the evolving structure of the estimation problem. 

\begin{figure}[t]
    \centering
    \includegraphics[width=0.9\linewidth]{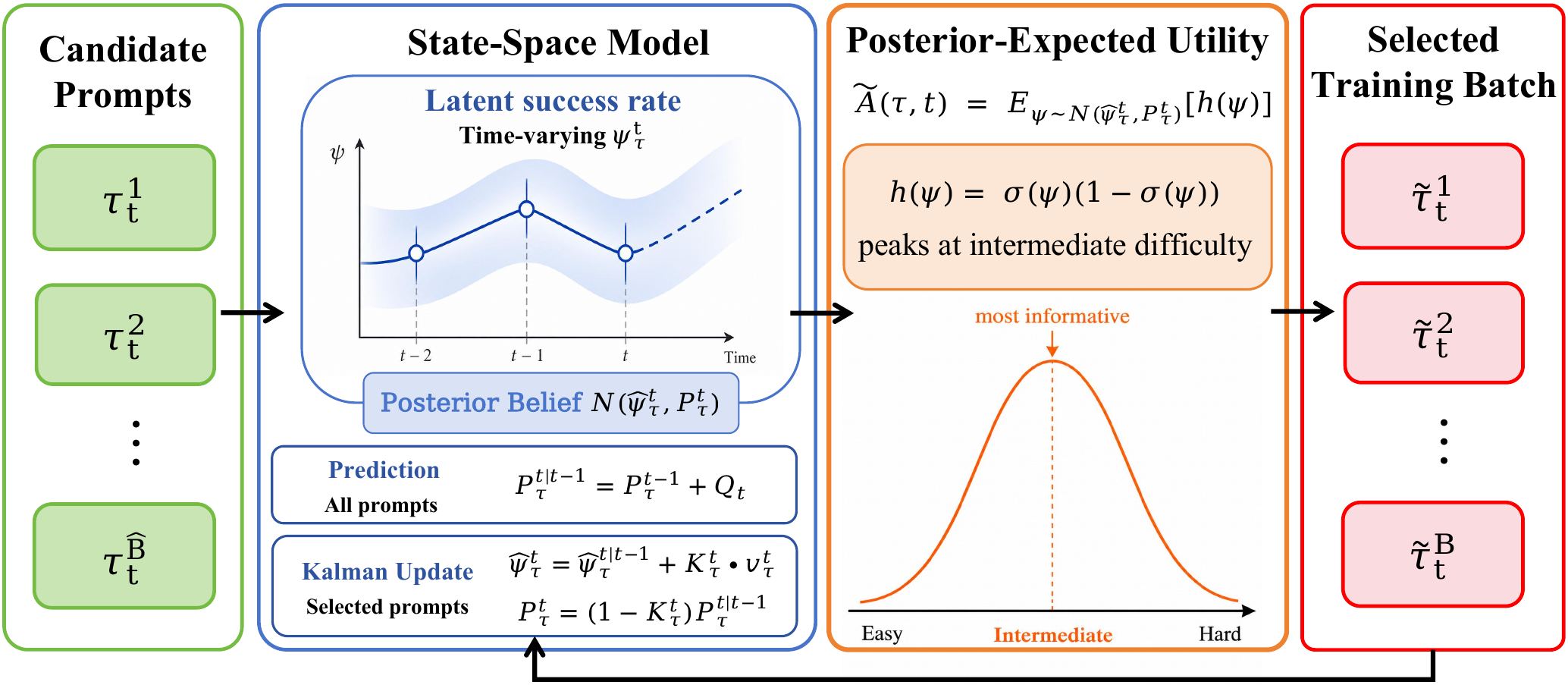}
    \caption{\textbf{Framework Overview of KGPS.} At each training step $t$, KGPS maintains a Gaussian posterior $\mathcal{N}(\hat\psi_\tau^t, P_\tau^t)$ over the latent logit-space difficulty of every candidate prompt via a Kalman filter.
    Prompts selected in the previous step receive a full Kalman update from their rollout observations; unselected prompts undergo only the prediction step, with variance inflated by $Q_t$ to reflect policy-induced uncertainty growth.
    The updated posteriors drive prompt selection through the posterior-expected utility $\tilde{A}(\tau, t) = \mathbb{E}_{\psi \sim \mathcal{N}(\hat\psi_\tau^t, P_\tau^t)}[h(\psi)]$, where $h(\psi) = \sigma(\psi)(1-\sigma(\psi))$ peaks at intermediate difficulty, ensuring the selected batch targets the most informative training signal while naturally revisiting long-neglected prompts. }
    \label{fig:main}
    \vspace{-15pt}
\end{figure}

\xu{To address this issue,} in this paper, we propose \emph{\textbf{K}alman-\textbf{G}uided \textbf{P}rompt \textbf{S}election} (\emph{\textbf{KGPS}}), a principled framework for online prompt selection that reframes difficulty tracking as a dynamic state estimation problem, \xu{as illustrated in Fig.~\ref{fig:main}.} 
Rather than modeling each prompt's success rate as a stationary latent parameter, KGPS adopts a linear-Gaussian state-space formulation in logit space, where the latent success rate of each prompt evolves as a random walk whose process noise is explicitly coupled to the magnitude of the policy update at each training step. 
This design ensures that the posterior uncertainty of every prompt, including those not selected for training, is continuously inflated to reflect the current state of the policy.
Building on this formulation, we derive a Kalman filter that maintains a Gaussian belief over each prompt's logit-transformed success rate, propagating uncertainty in a theoretically grounded manner. 
Prompt selection is then performed by maximizing a posterior-expected training utility score that prioritizes prompts at intermediate difficulty under the current policy, while naturally promoting exploration of prompts with accumulated uncertainty via posterior variance inflation. 
\xu{Importantly, KGPS operates on the same rollout outcomes already generated for policy training and requires no additional rollouts for prompt selection.}
% requiring no additional rollouts beyond standard policy training.
As illustrated in Fig.~\ref{fig:head}, KGPS substantially reduces success rate estimation error \xu{(red curve in Fig.~\ref{fig:head} (b))} and achieves superior performance under an equivalent rollout budget \xu{(red curve in Fig.~\ref{fig:head} (a))} compared to MoPPS \citep{qu2025promptdifficultyonlinepredicted}, a representative prediction-based method that relies on static posterior approximation, empirically validating the advantage of dynamic state estimation.
The primary contributions of this work are three-fold:
\begin{itemize}
    \item We reformulate prompt difficulty tracking as a dynamic state estimation 
    problem, modeling each prompt's latent success rate via a linear-Gaussian 
    state-space model with process noise coupled to the policy update magnitude.

    \item We derive a Kalman filter for online difficulty tracking and propose a 
    posterior-expected training utility score that naturally balances exploitation of 
    current estimates with exploration of uncertain prompts, requiring no hand-tuned 
    decay schedule.

    \item Extensive experiments across diverse reasoning tasks and RL algorithms demonstrate that KGPS 
    consistently outperforms existing baselines in both accuracy and rollout 
    efficiency with negligible computational overhead.
\end{itemize}

% \vspace{-10pt}

\section{Related Work}

% \paragraph{RL Finetuning of LLMs.}
% Reinforcement learning has become a dominant paradigm for enhancing the capabilities of large language models. Reinforcement Learning with Human Feedback (RLHF)~\citep{dong2024rlhf, dai2023saferlhf, zheng2023secretsrlhf} has demonstrated remarkable success in improving instruction following and ensuring safety alignment. More recently, Reinforcement Learning with Verifiable Rewards (RLVR)~\citep{jaech2024openai, guo2025deepseek, team2025kimi, tinyzero} has significantly advanced reasoning capabilities in structured domains such as mathematics and code generation, where correctness can be automatically verified. On the algorithmic side, PPO~\citep{schulman2017proximal} remains a widely adopted backbone, while GRPO~\citep{shao2024deepseekmath, deepseekai2025deepseekr1incentivizingreasoningcapability} offers a computationally lighter alternative by eliminating the value network and estimating advantages through group normalization. A growing body of work further improves these algorithms in terms of training stability, sample efficiency, and scalability~\citep{yu2025dapo, hu2025reinforce++, yue2025vapo, kazemnejad2024vineppo, sheng2024hybridflow}.

\paragraph{RL Finetuning of LLMs.}
Reinforcement learning has become a central paradigm for improving large language models, from alignment-oriented RLHF~\citep{dong2024rlhf, dai2023saferlhf, zheng2023secretsrlhf} to reasoning-oriented RLVR~\citep{jaech2024openai, guo2025deepseek, team2025kimi, tinyzero}. PPO~\citep{schulman2017proximal} remains a standard backbone. In the meantime, GRPO~\citep{shao2024deepseekmath, deepseekai2025deepseekr1incentivizingreasoningcapability} offers a lighter alternative. Recent work further improves RL finetuning in terms of stability, sample efficiency, and scalability~\citep{yu2025dapo, hu2025reinforce++, yue2025vapo, kazemnejad2024vineppo, sheng2024hybridflow}.

\paragraph{Prompt Selection for RL Finetuning.}
Data curation has emerged as a promising strategy for improving RL finetuning efficiency. 
\textit{Offline} methods select prompts prior to training based on static criteria such as difficulty or diversity~\citep{ye2025limo, li2025limr, hu2025open, yang2024qwen2math, fatemi2025concise}, but cannot adapt to the model's evolving capabilities. 
\textit{Online} methods address this by dynamically curating training batches, either filtering prompts with degenerate success rates~\citep{yu2025dapo, liu2025prorl, cui2025process, meng2025mm} or prioritizing intermediate-difficulty prompts to maximize gradient informativeness~\citep{bae2025online, chen2025self}, yet typically require additional rollouts over a large candidate pool, introducing substantial computational overhead. 
MoPPS~\citep{qu2025promptdifficultyonlinepredicted} mitigates this cost via inference-free prompt selection using Beta--Bernoulli posteriors with Thompson Sampling, but its static formulation assumes time-invariant prompt difficulty, failing to account for the non-stationary evolution induced by a continuously improving policy.

% \paragraph{Prompt Selection for RL Finetuning.}
% Data curation has emerged as a promising strategy for improving RL finetuning efficiency. Offline methods rely on static criteria such as difficulty or diversity~\citep{ye2025limo, li2025limr, hu2025open, yang2024qwen2math, fatemi2025concise}, while online methods adapt prompt selection during training~\citep{yu2025dapo, liu2025prorl, cui2025process, meng2025mm, bae2025online, chen2025self}, often at the cost of additional rollout overhead. MoPPS~\citep{qu2025promptdifficultyonlinepredicted} avoids this overhead via inference-free prompt selection with Beta--Bernoulli posteriors and Thompson Sampling, but its static formulation cannot capture the non-stationary evolution of prompt difficulty under an improving policy.

Our work is most closely related to these online prediction-based methods, but \textbf{differs} in one key aspect: instead of static parameter estimation with heuristic forgetting, we formulate prompt difficulty tracking as a dynamic state estimation problem. This allows uncertainty to evolve with policy updates and yields more accurate prompt difficulty estimates without requiring additional rollouts.

\section{Methodology}
\label{sec:method}

This section presents \textbf{KGPS} (\textbf{K}alman-\textbf{G}uided \textbf{P}rompt \textbf{S}election), as shown in Figure~\ref{fig:main}, \xu{our framework for addressing the non-stationarity problem identified in Sec.~\ref{sec:preliminary}. 
The key idea is to replace static difficulty estimation with dynamic state tracking. Concretely, KGPS proceeds in three stages: }
(i)~modeling each prompt's time-varying success rate as a latent state in a linear-Gaussian state-space model (Sec.~\ref{sec:ssm}); 
(ii)~applying a Kalman filter to propagate a Gaussian posterior over every prompt's logit success rate as the policy evolves (Sec.~\ref{sec:kalman}); 
and (iii)~selecting prompts by maximizing a posterior-expected training utility score derived from this posterior (Sec.~\ref{sec:selection}).

\subsection{Preliminary}
\label{sec:preliminary}

\noindent\textbf{Setup and notation.}
Let $\mathcal{T} = \{\tau_i\}_{i=1}^{N}$ denote the prompt pool of size $N$, $\pi_{\theta_t}$ is the policy at training step $t$, and $\mathcal{T}^t_B \subset \mathcal{T}$ denotes the selected batch of size $B$.
For each selected prompt $\tau \in \mathcal{T}^t_B$, the policy generates $k$ independent rollouts, each scored via a binary reward following a Bernoulli distribution~\citep{qu2025promptdifficultyonlinepredicted}:
\begin{equation}
    r_\tau^{t,j} \;\sim\; \mathrm{Bernoulli}(\phi_\tau^t),
    \qquad j = 1,\ldots,k,
    \label{eq:reward}
\end{equation}
where $\phi_\tau^t \in [0,1]$ is the \emph{latent success rate} of prompt $\tau$ under $\pi_{\theta_t}$, serving as a surrogate for its difficulty.
The empirical success count is $s_\tau^t = \sum_{j=1}^k r_\tau^{t,j}$, so that
$s_\tau^t \mid \phi_\tau^t \;\sim\; \mathrm{Binomial}(k,\, \phi_\tau^t)$.

\noindent\textbf{Bayesian online prompt selection.}
Existing methods~\citep{qu2025promptdifficultyonlinepredicted} model each prompt as an arm in a Bernoulli bandit and maintain a Beta--Bernoulli conjugate posterior over $\phi_\tau^t$.
Initialized with a uniform prior $\phi_\tau^0 \sim \mathrm{Beta}(1, 1)$, the posterior admits a closed-form recursive update:
\begin{equation}
    \phi_\tau^t \mid \mathcal{H}_t
    \;\sim\;
    \mathrm{Beta}(\alpha_\tau^t,\,\beta_\tau^t),
    \qquad
    \alpha_\tau^{t+1} = \lambda\,\alpha_\tau^t + s_\tau^t,
    \quad
    \beta_\tau^{t+1}  = \lambda\,\beta_\tau^t + k - s_\tau^t,
    \label{eq:beta-update}
\end{equation}
where $\mathcal{H}_t = \{\mathcal{T}_i^B, \mathcal{R}_i^B\}_{i=0}^{t}$ is the optimization history and $\lambda \in (0,1)$ is a decay factor introduced to discount stale observations.
Prompts are then selected by preferring posterior means $\hat{\phi}_\tau^t = \alpha_\tau^t / (\alpha_\tau^t + \beta_\tau^t)$ near the target $\phi^* \approx 0.5$~\citep{bae2025online, chen2025self}, where gradient informativeness is highest.

\vspace{-10pt}

\paragraph{Limitation: the non-stationarity problem.}
\zhd{
The Beta--Bernoulli framework implicitly treats $\phi_\tau^t$ as a stationary latent parameter; while the decay factor $\lambda$ down-weights stale observations, it provides no principled mechanism for predicting how $\phi_\tau^t$ evolves between selections. 
As shown in Figure~\ref{fig:head} (b), this leads to persistently high estimation error throughout training. 
In early training, the Beta posterior remains diffuse despite warmup initialization. As MoPPS~\citep{qu2025promptdifficultyonlinepredicted} acknowledge, estimation accuracy improves only as the pseudo-count $C_\tau^t = \alpha_\tau^t + \beta_\tau^t$ grows, a condition that is rarely met given the limited rollouts available per prompt in practice.
Thompson Sampling further compounds the issue by introducing additional stochastic variance into the predicted success rates. 
In later training, the fixed $\lambda$ discounts observations at a constant rate regardless of policy update magnitude, failing to track the non-stationarity difficulty drift induced by a continuously improving policy $\pi_{\theta_t}$. 
Resolving this model--reality mismatch requires reframing prompt difficulty tracking from static parameter estimation to \emph{dynamic state estimation}, which is the central objective of KGPS.
}

% ----------------------------------------------------------
\subsection{State-Space Formulation in Logit Space}
\label{sec:ssm}
% ----------------------------------------------------------

We treat each prompt's success rate as a \textbf{time-varying latent
state} and adopt a linear-Gaussian state-space model (SSM) to track it.
This choice is motivated by two considerations:
(i)~SSMs provide a principled, closed-form mechanism for propagating
uncertainty under non-stationarity, without recourse to hand-tuned decay
schedules; and
(ii)~linear-Gaussian \xu{SSMs admit the Kalman filter under Gaussian approximation}, producing a well-calibrated posterior that Sec.~\ref{sec:selection} subsequently uses to score and select prompts.
Let $\phi_\tau^t \in [0, 1]$ denote the expected success rate of prompt $\tau$ under the current policy $\pi_{\theta_t}$ at step $t$.
We work in logit space with $\psi_\tau^t = \mathrm{logit}(\phi_\tau^t) \in (-\infty, +\infty)$ as the latent state, satisfying the unconstrained domain requirement of the linear-Gaussian framework.

\noindent\textbf{State evolution.}
We model the latent logit $\psi_\tau^t$ as a random walk:

\begin{equation}
  \psi_\tau^t = \psi_\tau^{t-1} + w_\tau^t, \qquad
  w_\tau^t \sim \mathcal{N}(0,\, Q_t),
  \label{eq:state-evolution}
\end{equation}
where the process noise variance is tied to the magnitude of the preceding policy update:
\begin{equation}
  Q_t = \gamma \cdot \|\theta_{t} - \theta_{t-1}\|_2,
  \label{eq:Q}
\end{equation}
with $\gamma > 0$ a scalar hyperparameter.
This coupling reflects a natural intuition: a larger policy step is likely to shift prompt difficulty more, warranting greater state uncertainty.
Note that $Q_t$ depends only on the \emph{previous} parameter update, which is available at the start of step $t$ before any new rollouts are collected. 
This ensures the prediction step (Sec.~\ref{sec:kalman}) is executable without circular dependency.

\paragraph{Observation model.}
At step $t$, if prompt $\tau$ is selected with $k$ rollouts yielding $s_\tau^t$ successes, the empirical success rate $\hat\phi_\tau^t = s_\tau^t / k$ follows a scaled Binomial distribution (as defined in Sec.~\ref{sec:preliminary}), which is incompatible with the Gaussian observation requirement of the linear-Gaussian framework. 
We therefore apply the delta method to the logit-transformed $\hat\phi_\tau^t$, which propagates the known variance $\mathrm{Var}(\hat\phi_\tau^t) = \phi_\tau^t(1-\phi_\tau^t)/k$ through the logit transform to yield a tractable Gaussian approximation in logit space (derivation in Appendix~\ref{app:detla}):

\begin{equation}
  \mathrm{logit}(\hat\phi_\tau^t) = \psi_\tau^t + \varepsilon_\tau^t,
  \qquad \varepsilon_\tau^t \sim \mathcal{N}(0,\, R_\tau^t).
  \label{eq:obs-model}
\end{equation}

Since the observation noise variance $R_\tau^t = 1/[k\phi_\tau^t(1-\phi_\tau^t)]$ depends on the unknown $\phi_\tau^t$, we substitute the current posterior mean $\hat\psi_\tau^{t}$ at the start of step $t$ before rollouts are collected, produced by the Kalman filter at step $t-1$ (see Sec.~\ref{sec:kalman}). $\hat\psi_\tau^{t}$ is in fact a plug-in estimator that models the empirical success rate $\hat\phi_\tau^t$ as:
\begin{equation}
    \tilde\phi_\tau^t = {clip}(\sigma(\hat\psi_\tau^{t}),\,\delta,\,1-\delta), 
\end{equation}
where $\delta = 1/(2k)$ ensures $R_\tau^t$ remains bounded above, and $\sigma(\cdot)$ denotes the sigmoid function mapping from logit space back to $[0,1]$. Therefore, the observation noise variance is formulated as:
\begin{equation}
  R_\tau^t = \frac{1}{k\,\tilde\phi_\tau^t(1-\tilde\phi_\tau^t)}.
  \label{eq:R}
\end{equation}

% ----------------------------------------------------------
\subsection{Kalman Filtering for Posterior Maintenance}
\label{sec:kalman}
% ----------------------------------------------------------

With the SSM of Eqs.~\eqref{eq:state-evolution}--\eqref{eq:R} in hand, we apply a Kalman filter (KF) to maintain a Gaussian posterior $\mathcal{N}(\hat\psi_\tau^t, P_\tau^t)$ over the latent logit state $\psi_\tau^t$ for every prompt $\tau \in \mathcal{T}$, where $\hat\psi_\tau^t$ is the posterior mean estimate and $P_\tau^t$ is the posterior variance.
The filter's responsibility is to maintain well-calibrated posteriors as the policy evolves, with the resulting pair $(\hat\psi_\tau^t, P_\tau^t)$ subsequently serving as the input to the training utility score and prompt selection in Sec.~\ref{sec:selection}.
Since both the state transition and observation maps are linear with identity coefficients, the KF reduces to simple scalar recursions.

\paragraph{Prediction step (all prompts, every step).}
At each training step $t$, we inflate the posterior variance of
\emph{all} prompts to account for the latest policy update, regardless
of whether they were selected:
\begin{equation}
  P_\tau^{t\mid t-1} = P_\tau^{t-1} + Q_t,
  \qquad \forall\,\tau\in\mathcal{T}.
  \label{eq:predict}
\end{equation}
Because $Q_t > 0$, the predicted variance of every prompt grows at each step, even for those not recently evaluated.
This monotonic growth is what allows the learnability score in Sec.~\ref{sec:selection} to naturally favour prompts that have been neglected: their inflated $P_\tau^{t|t-1}$ raises their score deterministically until they are eventually selected.

\paragraph{Update step (selected prompts only).}
For each $\tau \in \mathcal{T}_{t-1}^B$, {\em i.e.}, prompts selected at step $t-1$, the filter incorporates the logit-space observation via the standard KF equations:
\begin{equation}
  K_\tau^t = \frac{P_\tau^{t\mid t-1}}{P_\tau^{t\mid t-1} + R_\tau^{t-1}}, \quad
  \hat\psi_\tau^t = \hat\psi_\tau^{t\mid t-1} + K_\tau^t\cdot\nu_\tau^t, \quad
  P_\tau^t = \bigl(1 - K_\tau^t \bigr)\,P_\tau^{t\mid t-1}.
  \label{eq:kalman-update}
\end{equation}

where $\hat\psi_\tau^{t\mid t-1} = \hat\psi_\tau^{t-1}$ is the one-step predicted mean and $\nu_\tau^t = \mathrm{logit}(\hat\phi_\tau^{t-1}) - \hat\psi_\tau^{t\mid t-1}$ is the innovation.
For prompts not selected at step $t-1$, the state estimate is unchanged and the variance retains its predicted value $P_\tau^{t\mid t-1}$.
The Kalman gain $K_\tau^t \in (0,1]$ adaptively weights the new observation against the prior belief. 
When the rollout is near-degenerate, the clipping in Eq.~\eqref{eq:R} ensures $R_\tau^{t-1}$ remains bounded, so the filter retains a non-negligible update even for extreme observations. 
Conversely, when the prior is diffuse ($P_\tau^{t\mid t-1}$ large) and the observation is reliable (moderate $\hat\phi_\tau^{t-1}$, small $R_\tau^{t-1}$), $K_\tau^t \to 1$ and the posterior mean shifts aggressively toward the new data. 
In both cases, the updated $(\hat\psi_\tau^t, P_\tau^t)$ faithfully encodes the filter's current state of knowledge about each prompt's difficulty, which is precisely what Sec.~\ref{sec:selection} requires.

% ----------------------------------------------------------
\subsection{Prompt Selection via Posterior-Expected Utility}
\label{sec:selection}
% ----------------------------------------------------------

At each training step $t$, the Kalman filter delivers a posterior $\mathcal{N}(\hat\psi_\tau^t, P_\tau^t)$ for each prompt.
We now show how to convert this posterior into a scalar training utility score and use it to select the most informative batch for the next training step.

\paragraph{From point estimates to posterior expectations.}
Following prior work~\citep{qu2025promptdifficultyonlinepredicted,bae2025online, chen2025self}, training efficiency is maximized by selecting prompts at intermediate difficulty, where the current policy is neither saturated nor overwhelmed. 
This motivates scoring each prompt by its estimated training utility $h(\hat\psi_\tau^t)$, where $h(\psi)=\sigma(\psi)(1-\sigma(\psi))$ peaks at intermediate success rates and vanishes at the extremes.
This criterion, however, ignores the posterior variance $P_\tau^t$: a prompt with a near-degenerate success rate is persistently excluded from future selection even when its true difficulty under the current policy remains unknown.
We resolve this by replacing the point evaluation with its expectation
under the full Gaussian posterior maintained by the filter:
\begin{equation}
  \tilde{A}(\tau,\,t)
  =
  \mathbb{E}_{\psi\,\sim\,\mathcal{N}(\hat\psi_\tau^t,\,P_\tau^t)}
  \!\bigl[h(\psi)\bigr].
  \label{eq:score-posterior}
\end{equation}
Because $h(\psi)$ is nonlinear and bounded, $\tilde{A}(\tau,t)$ is greater than $h(\hat\psi_\tau^t)$ whenever $P_\tau^t > 0$: integrating over the posterior spreads probability mass toward the intermediate-difficulty region where $h$ is large, so any prompt with accumulated uncertainty retains a non-negligible score regardless of where its mean currently sits.
Crucially, no stochastic sampling is needed: the prediction step in Eq.~\eqref{eq:predict} continuously inflates $P_\tau^t$ for unobserved prompts, which raises $\tilde{A}(\tau,t)$ deterministically until the prompt is revisited.

\paragraph{Efficient computation via Gauss--Hermite quadrature.}
The expectation in Eq.~\eqref{eq:score-posterior} has no closed form.
Since $h(\psi)$ is a smooth, bounded function of a scalar Gaussian variable, we evaluate it efficiently via five-point Gauss--Hermite quadrature (detailed in Appendix~\ref{app:gh}).
Substituting $\psi = \hat\psi_\tau^t + \sqrt{2P_\tau^t}\,x$ yields:
\begin{equation}
  \tilde{A}(\tau,\,t)
  \approx
  \frac{1}{\sqrt{\pi}}
  \sum_{i=1}^{5} w_i\,
  h\!\left(\hat\psi_\tau^t + \sqrt{2P_\tau^t}\,x_i\right),
  \label{eq:gh}
\end{equation}
where the node--weight pairs $(x_i,w_i)\in\{(0,\;0.9453),\;(\pm0.9586,\;0.3936),\; (\pm2.0202,\;0.0200)\}$ are fixed constants requiring no tuning.
The per-prompt cost is five scalar evaluations of $h(\psi)$, negligible relative to LLM rollout overhead.

\paragraph{Prompt selection.}
At each step $t$, we compute $\tilde{A}(\tau,t)$ for every
$\tau\in\mathcal{T}$ and select the training batch by greedy
maximization:
\begin{equation}
  \mathcal{T}^t_B
  = \operatorname{Top-}B
    \bigl\{\tau\in\mathcal{T}
    \;\big|\;
    \tilde{A}(\tau,\,t)\bigr\}.
  \label{eq:topB}
\end{equation}
Before selection begins, we run a single warmup epoch in which prompts are sampled uniformly to obtain an initial empirical success rate $\hat\phi_\tau^0$ for every $\tau \in \mathcal{T}$. 
The Kalman filter is then initialized with $\hat\psi_\tau^0 = \mathrm{logit}(\hat\phi_\tau^0)$, providing a data-driven starting point for the posterior mean rather than a uninformative init prior. 
The full KGPS pipeline is summarized in Algorithm~\ref{alg:kgps} in the Appendix.

\section{Experiment}
\label{sec:experiments}
% This section conducts experiments to evaluate whether KGPS can accurately track prompt difficulty online, improve rollout efficiency, and boost final performance across diverse reasoning tasks and RL algorithms. We further analyze the contribution of each component through ablation studies.
% This section evaluates KGPS's online difficulty tracking, rollout efficiency, and final performance across reasoning tasks and RL algorithms, plus ablation studies of each component.

\begin{figure}[t]
    \centering
    \begin{minipage}{0.32\linewidth}
        \centering
        \includegraphics[width=\linewidth, height=0.16\textheight]{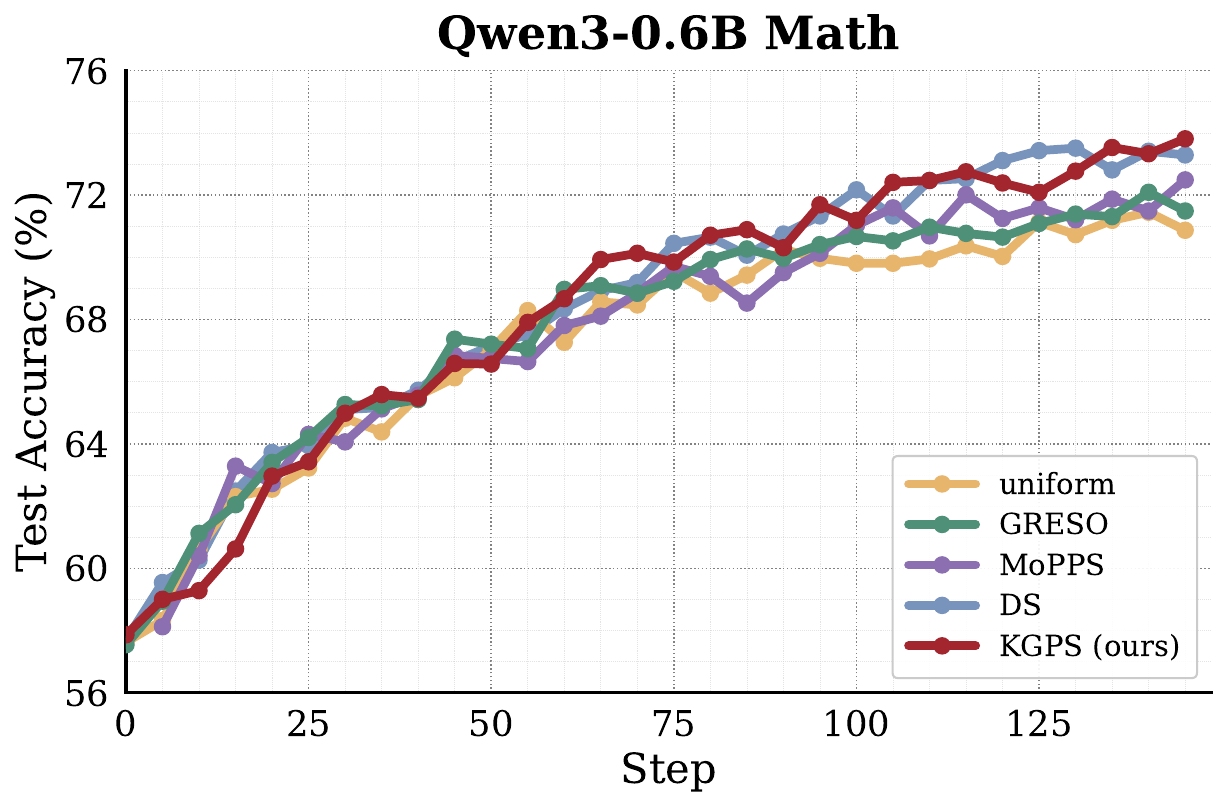}
    \end{minipage}
    \hfill
    \begin{minipage}{0.32\linewidth}
        \centering
        \includegraphics[width=\linewidth, height=0.16\textheight]{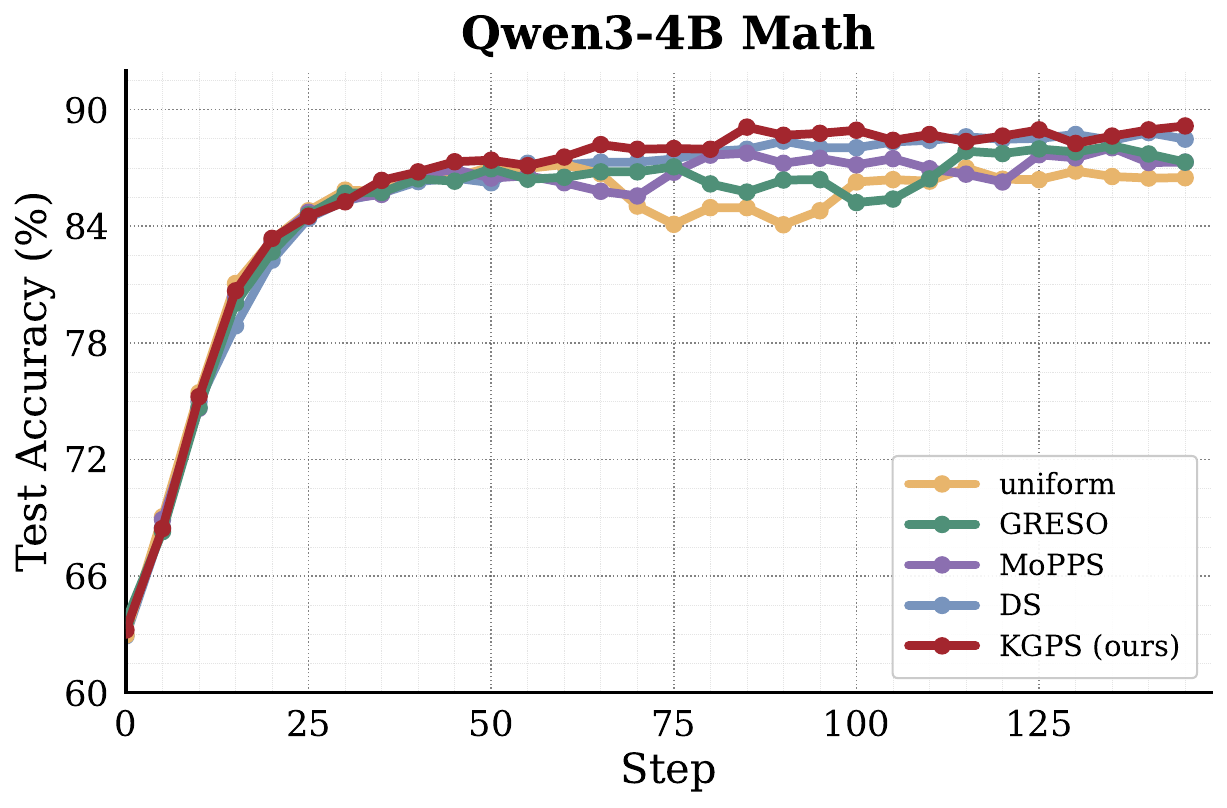}
    \end{minipage}
    \hfill
    \begin{minipage}{0.32\linewidth}
        \centering
        \includegraphics[width=\linewidth, height=0.16\textheight]{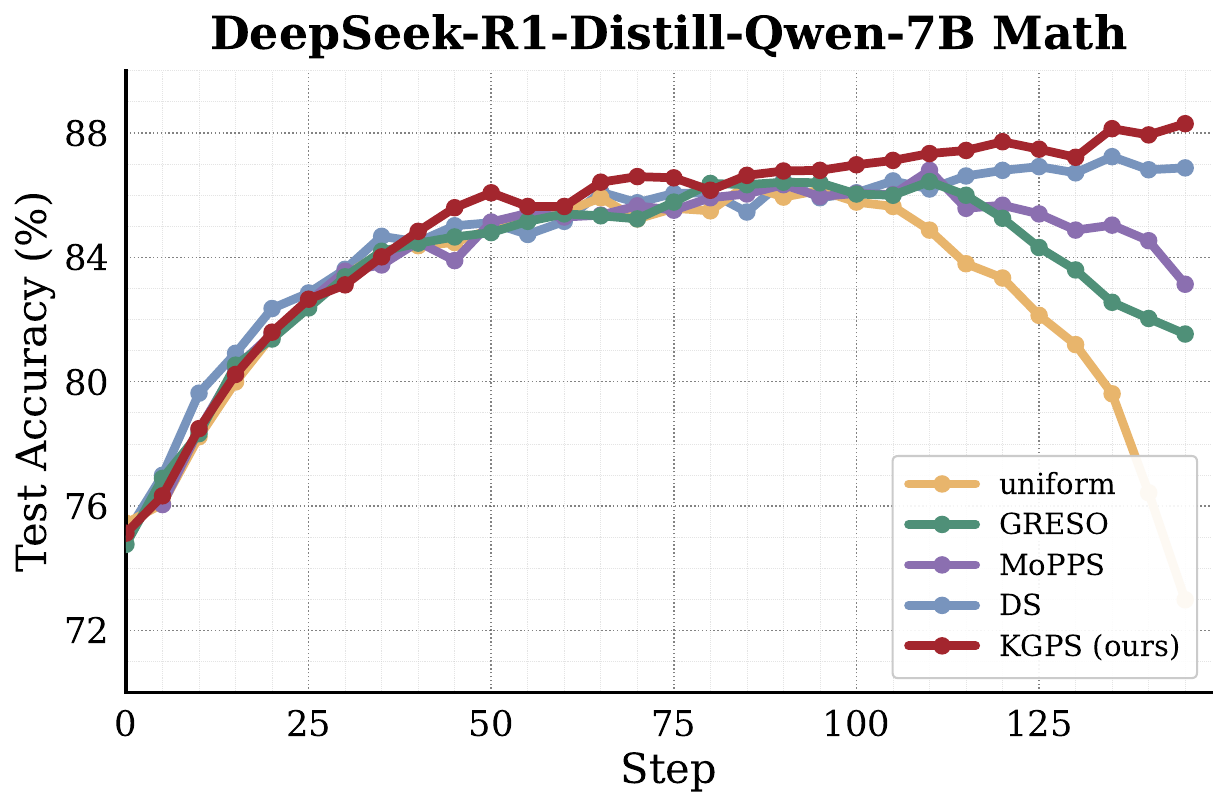}
    \end{minipage}
    \caption{Test accuracy on the Math benchmark across three model scales under different data selection strategies. KGPS (ours) consistently achieves top performance.}
    \label{fig:Math}
\end{figure}

\begin{figure}[t]
    \centering
    \begin{minipage}{0.32\linewidth}
        \centering
        \includegraphics[width=\linewidth, height=0.16\textheight]{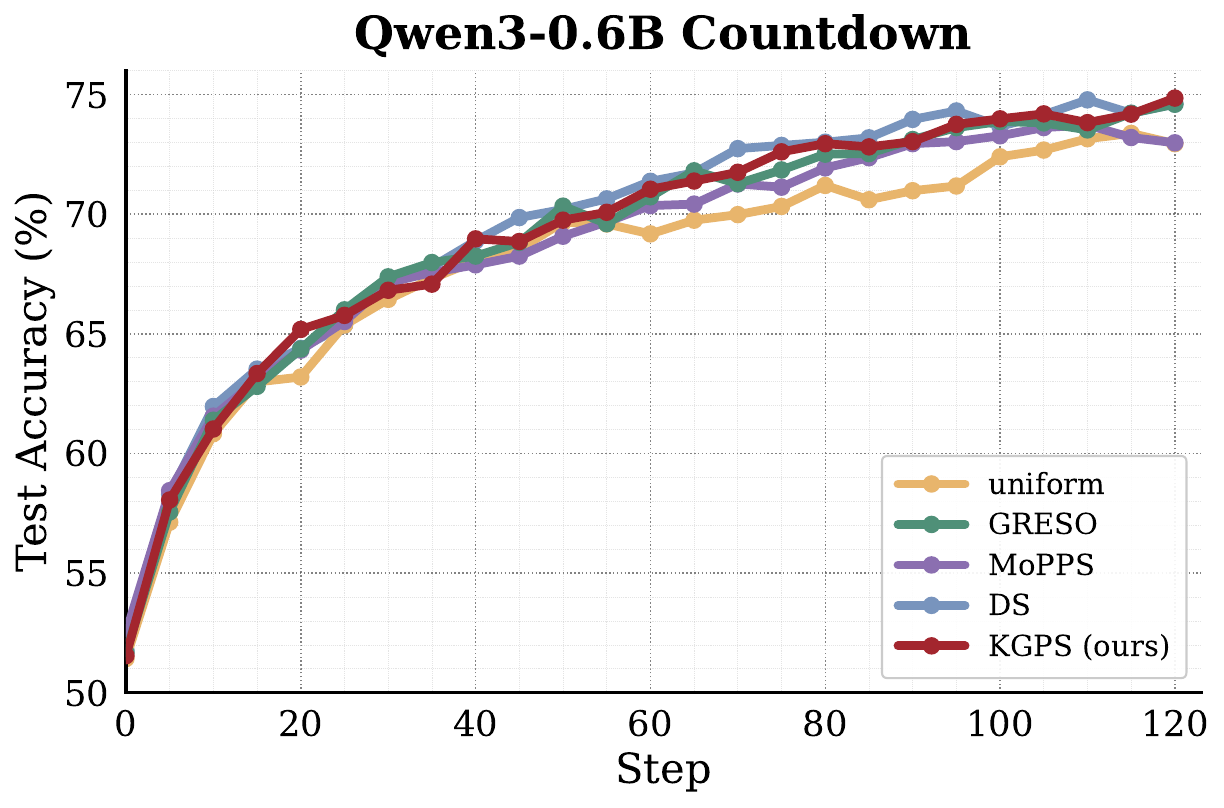}
    \end{minipage}
    \hfill
    \begin{minipage}{0.32\linewidth}
        \centering
        \includegraphics[width=\linewidth, height=0.16\textheight]{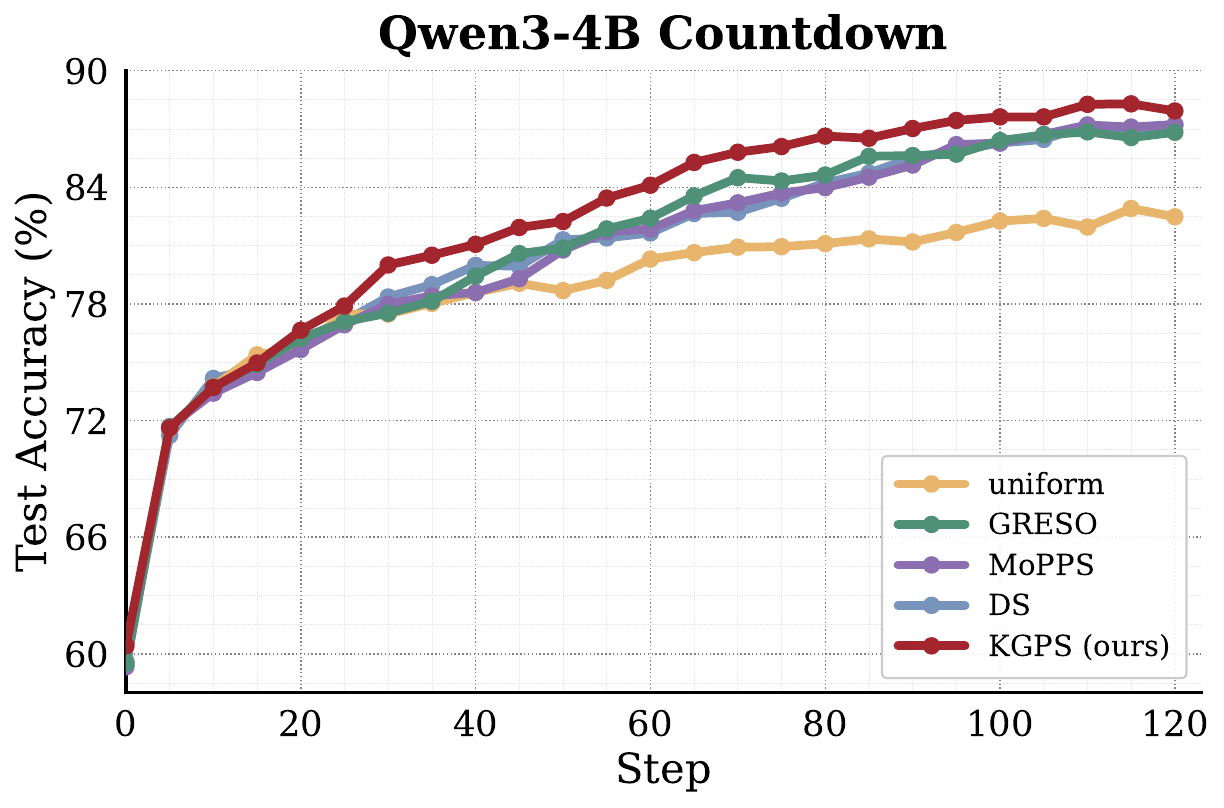}
    \end{minipage}
    \hfill
    \begin{minipage}{0.32\linewidth}
        \centering
        \includegraphics[width=\linewidth, height=0.16\textheight]{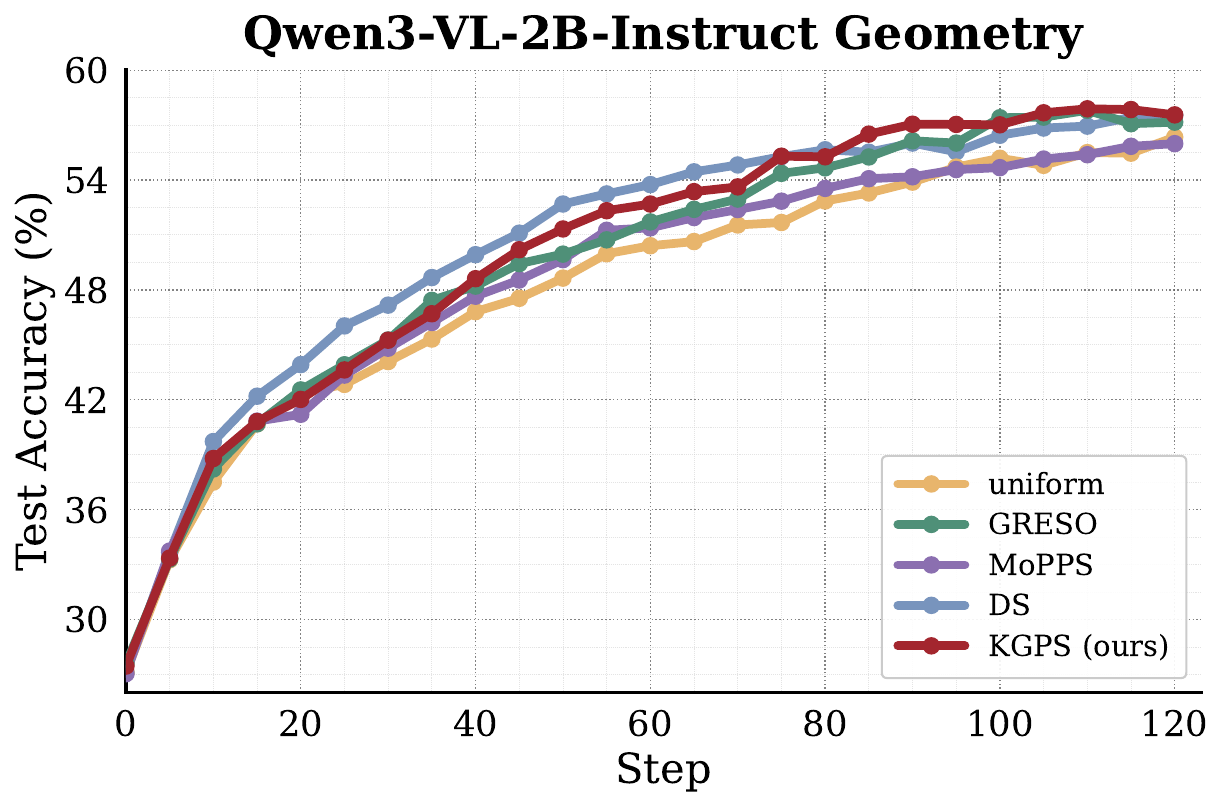}
    \end{minipage}
    \caption{Test accuracy on Countdown and Geometry benchmarks, where KGPS (ours) demonstrates consistent advantages over all baseline data selection strategies across varying model scales.}
    \label{fig:countdown}
    \vspace{-10pt}
\end{figure}

\begin{table}[t]
\renewcommand{\arraystretch}{1.2}
  \centering
  \caption{Evaluation across mathematics benchmarks. `+' represents 
  finetuning with the method. \textbf{Bold} indicates the best result 
  in each column.}
  \resizebox{\linewidth}{!}{
    \begin{tabular}{lccccccccc}
    \toprule
    \multicolumn{1}{c}{Method} & AIME24 & AIME25 & AMC23 & MATH500 & Minerva. & Olympiad. & Avg. $\uparrow$ & Rollouts$\downarrow$ & Runtime$\downarrow$\\
    \midrule
    \rowcolor{gray!20} \multicolumn{10}{c}{Qwen3-0.6B} \\
    +Uniform    & 4.79  & 11.67 & 38.63 & 69.35 & 16.61 & 33.51 & 29.09 & 296k & 10.87h \\
    +DS         & 3.54 & 15.62 & 41.27 & 70.88 & \textbf{17.58} & 35.28 & 30.70 & 1024k & 24.66h \\
    +GRESO      & 2.92  & 12.92 & 39.68 & 68.85 & 15.92 & 32.83 & 28.85 & 296k & 10.48h \\
    +MoPPS      & 3.13 & 14.37 & 40.96 & 70.78 & 16.89 & 34.06 & 30.03 & 296k & 10.48h \\
    +KGPS(Ours) & \textbf{5.42} & \textbf{15.83} & \textbf{43.67} & \textbf{72.29} & 17.51 & \textbf{35.45} & \textbf{31.70} & 296k & 10.74h \\
    \midrule
    \rowcolor{gray!20} \multicolumn{10}{c}{Qwen3-4B} \\
    +Uniform    & 24.38 & 29.68 & 66.87 & 84.76 & 30.65 & 50.88 & 47.87 & 296k & 17.21h \\
    +DS         & 26.25 & 32.92 & 71.46 & 87.88 & \textbf{31.73} & 54.48 & 50.79 & 2310k & 47.86h \\
    +GRESO      & 23.13 & 33.75 & 70.86 & 86.90 & 31.18 & 52.45 & 49.71 & 296k & 17.31h \\
    +MoPPS      & \textbf{27.08} & 32.50 & 71.61 & 85.39 & 31.25 & 52.89 & 50.12 & 296k & 16.29h \\
    +KGPS(Ours) & 26.46 & \textbf{36.04} & \textbf{73.27} & \textbf{88.04} & 31.55 & \textbf{54.63} & \textbf{51.67} & 296k & 16.69h \\
    \midrule
    \rowcolor{gray!20} \multicolumn{10}{c}{DeepSeek-R1-Distill-Qwen-7B} \\
    +Uniform    & 28.75 & 22.92 & 60.17 & 77.76 & 28.99 & 46.17 & 44.13 & 296k & 11.54h \\
    +DS         & \textbf{47.08} & 34.69 & 78.58 & 86.30 & 30.27 & 55.73 & 55.44 & 1744k & 36.53h \\
    +GRESO      & 32.40 & 25.94 & 66.00 & 80.35 & 30.88 & 49.19 & 47.46 & 296k & 11.39h \\
    +MoPPS      & 31.35 & 26.56 & 67.28 & 82.43 & \textbf{30.96} & 50.59 & 48.20 & 296k & 11.43h \\
    +KGPS(Ours) & 46.35 & \textbf{35.73} & \textbf{78.84} & \textbf{86.32} & 30.35 & \textbf{55.76} & \textbf{55.56} & 296k & 11.91h \\
    \bottomrule
    \end{tabular}
    }
  \vspace{-4pt}
  \label{tab:matheval}
\end{table}

\subsection{Experimental Setup}

\paragraph{Datasets and Models.}
% We evaluate KGPS on three reasoning benchmarks spanning distinct modalities.
% For \textbf{mathematics}, models are trained on MATH~\citep{hendrycksmath2021} and evaluated on six benchmarks: MATH500~\citep{hendrycksmath2021,lightman2023lets}, AIME 2024~\citep{aime2024}, AIME 2025~\citep{aime2025}, AMC 2023~\citep{amc2023}, Minerva Math~\citep{lewkowycz2022solving}, and OlympiadBench~\citep{he2024olympiadbench}.
% For \textbf{planning}, we use the Countdown Number Game~\citep{tinyzero}, training and evaluating on the CD-34 dataset.
% For \textbf{visual geometry}, we train and evaluate on Geometry3k~\citep{lu2021inter,geometry3k_dataset}, requiring joint visual perception and multi-step geometric reasoning.
% We evaluate across diverse backbones: Qwen3-\{0.6B,4B\}, R1-Distill-Qwen-7B, and Qwen3-8B for mathematics; Qwen3-\{0.6B,4B\} for planning; Qwen3-VL-2B-Instruct and Qwen2.5-VL-7B-Instruct for visual geometry.

We evaluate KGPS on three reasoning benchmarks across distinct modalities. For \textbf{mathematics}, models train on MATH~\citep{hendrycksmath2021} and test on MATH, MATH500, AIME 2024/2025, AMC 2023, Minerva Math, and OlympiadBench~\citep{hendrycksmath2021,lightman2023lets,aime2024,aime2025,amc2023,lewkowycz2022solving,he2024olympiadbench}. For \textbf{planning}, we use Countdown (CD-34)~\citep{tinyzero}. For \textbf{visual geometry}, we use Geometry3k~\citep{lu2021inter,geometry3k_dataset}. Backbones: Qwen3-{0.6B}, Qwen3-{4B}, Qwen3-8B, and DeepSeek-R1-Distill-Qwen-7B for math; Qwen3-{0.6B}, and Qwen3-{4B} for planning; Qwen3-VL-2B-Instruct and Qwen2.5-VL-7B-Instruct for geometry.
% \vspace{-8pt}

\paragraph{Training Protocol.}
% All models are fine-tuned on the \texttt{verl} framework~\cite{sheng2024hybridflow}, with $k$ rollouts per prompt per step.
% Test accuracy is reported as average pass@1 over 16 independent generations on held-out sets throughout training.
% Implementation details are in the Appendix \ref{app:impl}.

Models are fine-tuned on \texttt{verl}~\cite{sheng2024hybridflow} with $8$ rollouts per prompt per step. Test accuracy is reported as average pass@1 over 16 generations on held-out sets. See Appendix~\ref{app:impl} for details.
% \vspace{-8pt}

\paragraph{Baselines.}
% We compare against four baselines:
% \textbf{Uniform} samples prompts uniformly at random;
% \textbf{GRESO}~\citep{greso} filters prompts with near-degenerate success rates;
% \textbf{MoPPS}~\citep{qu2025promptdifficultyonlinepredicted} maintains Beta--Bernoulli posteriors with Thompson Sampling selection;
% \textbf{DS (Oracle)}~\citep{yu2025dapo} selects from a large oversampled pool based on observed rewards, serving as an oracle upper bound at substantially higher rollout cost.
% Our primary goal is to match or exceed DS accuracy within the same rollout budget as Uniform, demonstrating that principled dynamic state estimation can achieve oracle-quality selection without additional computational overhead.
We compare against four baselines:
\textbf{Uniform} (random sampling);
\textbf{GRESO}~\citep{greso} (filters near-degenerate prompts);
\textbf{MoPPS}~\citep{qu2025promptdifficultyonlinepredicted} (Beta-Bernoulli posteriors + Thompson Sampling);
\textbf{DS (Oracle)}~\citep{yu2025dapo} (oversamples and filters, higher rollout cost).
Our goal is to match or exceed DS accuracy under Uniform's budget, showing dynamic state estimation achieves oracle quality without extra cost.

\subsection{Main Results}

\paragraph{Mathematics.}
% Fig.~\ref{fig:Math} shows that KGPS maintains a consistently superior accuracy trajectory throughout training, matching or exceeding DS across all three model scales without requiring real-time evaluation feedback. 
% On R1-Distill-7B, competing methods exhibit notable accuracy degradation in later steps, while KGPS sustains steady improvement. 
% To assess out-of-distribution generalization, we further evaluate all methods on six held-out benchmarks.
% As shown in Tab.~\ref{tab:matheval}, KGPS achieves the highest average accuracy among all prompt selection methods across all three model scales.

Fig.~\ref{fig:Math} shows KGPS maintains a superior accuracy trajectory across all model scales, matching or exceeding DS without real-time feedback. On DeepSeek-R1-Distill-Qwen-7B, KGPS avoids degradation seen in other methods. Tab.~\ref{tab:matheval} confirms KGPS achieves the highest average accuracy among selection methods.
%
% \xu{Most notably, KGPS matches or slightly surpasses DS, an evaluation-based baseline with access to real-time rollout feedback, while consuming substantially fewer rollouts:} 71\%, 87\%, and 83\% fewer on Qwen3-0.6B, Qwen3-4B, and R1-Distill-7B, respectively, with average accuracy gains of +1.00, +0.88, and +0.12 points.
% Under the same rollout budget of 296k, KGPS outperforms MoPPS by +1.67, +1.55, and +7.36 points on the three models, demonstrating that dynamic posterior tracking yields consistent gains over static difficulty estimation.
% KGPS also surpasses Uniform and GRESO across all setups, confirming the benefit of uncertainty-aware prompt selection over non-adaptive strategies.
% 
\xu{Remarkably, KGPS matches or slightly surpasses DS, an evaluation-based baseline with access to real-time rollout feedback, while using 71\%, 87\%, and 83\% fewer rollouts on Qwen3-0.6B, Qwen3-4B, and DeepSeek-R1-Distill-Qwen-7B respectively, with average accuracy gains of 1.00, 0.88, and 0.12 points.} Under the same 296k budget, KGPS outperforms MoPPS by 1.67, 1.55, 7.36 points. KGPS also surpasses Uniform and GRESO across all setups, confirming the benefit of uncertainty-aware prompt selection over non-adaptive strategies.

\begin{figure}[t]
    \centering
    % 第一行：Math 3张图
    \begin{minipage}{0.32\linewidth}
        \centering
        \includegraphics[width=\linewidth, height=0.15\textheight]{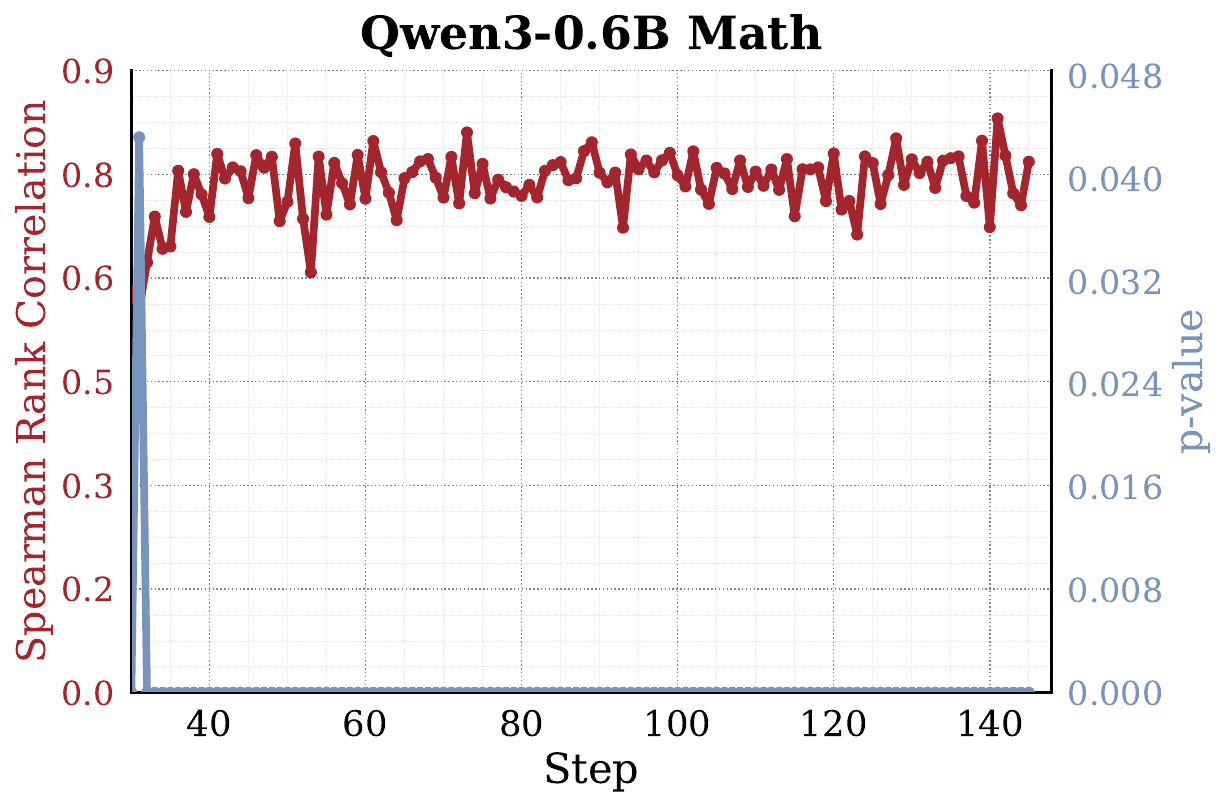}
    \end{minipage}
    \hfill
    \begin{minipage}{0.32\linewidth}
        \centering
        \includegraphics[width=\linewidth, height=0.15\textheight]{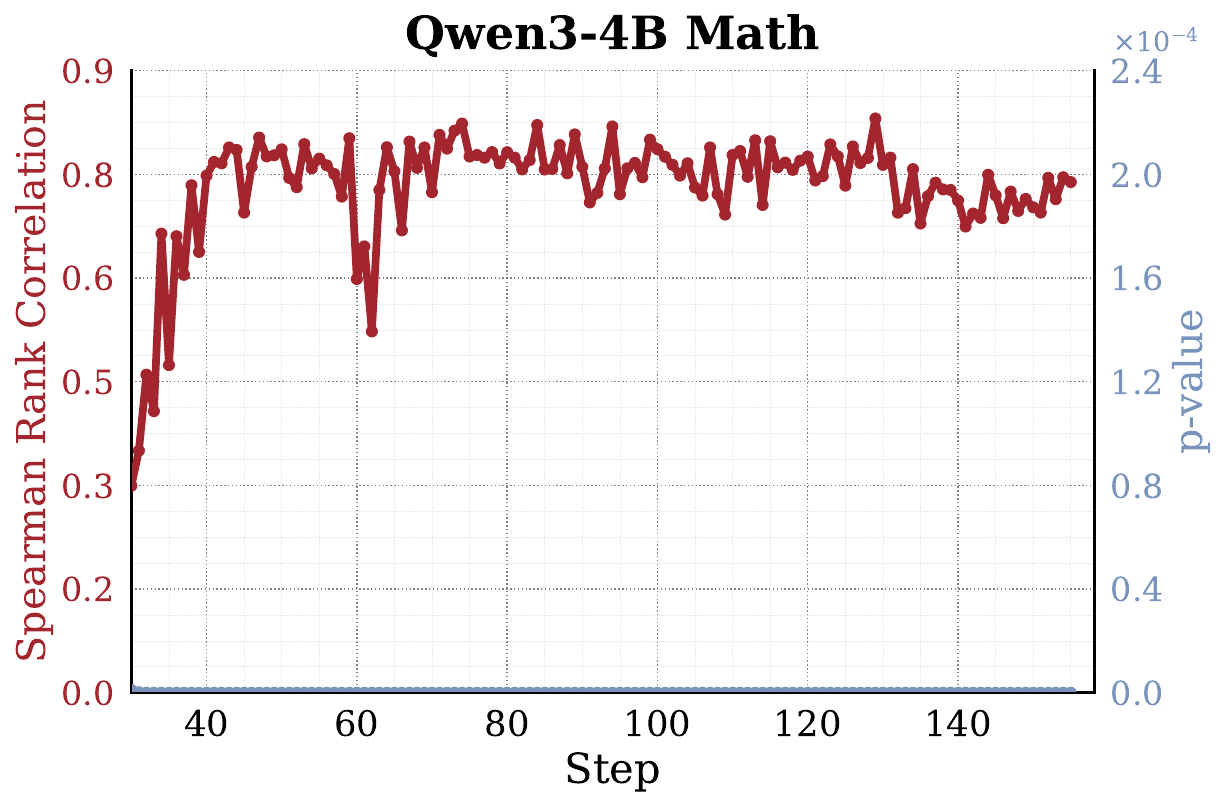}
    \end{minipage}
    \hfill
    \begin{minipage}{0.32\linewidth}
        \centering
        \includegraphics[width=\linewidth, height=0.15\textheight]{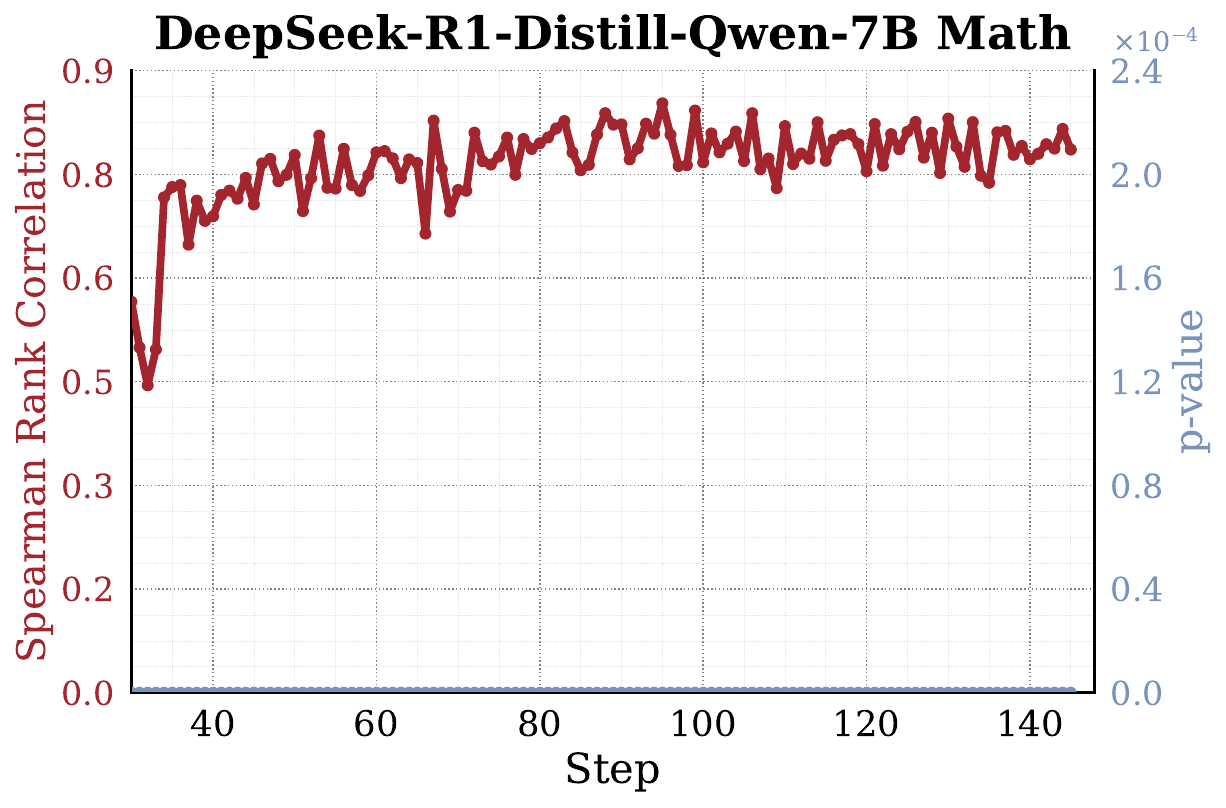}
    \end{minipage}

    \vspace{6pt} % 两行之间的间距，可调整

    % 第二行：Countdown + Geometry 3张图
    \begin{minipage}{0.32\linewidth}
        \centering
        \includegraphics[width=\linewidth, height=0.16\textheight]{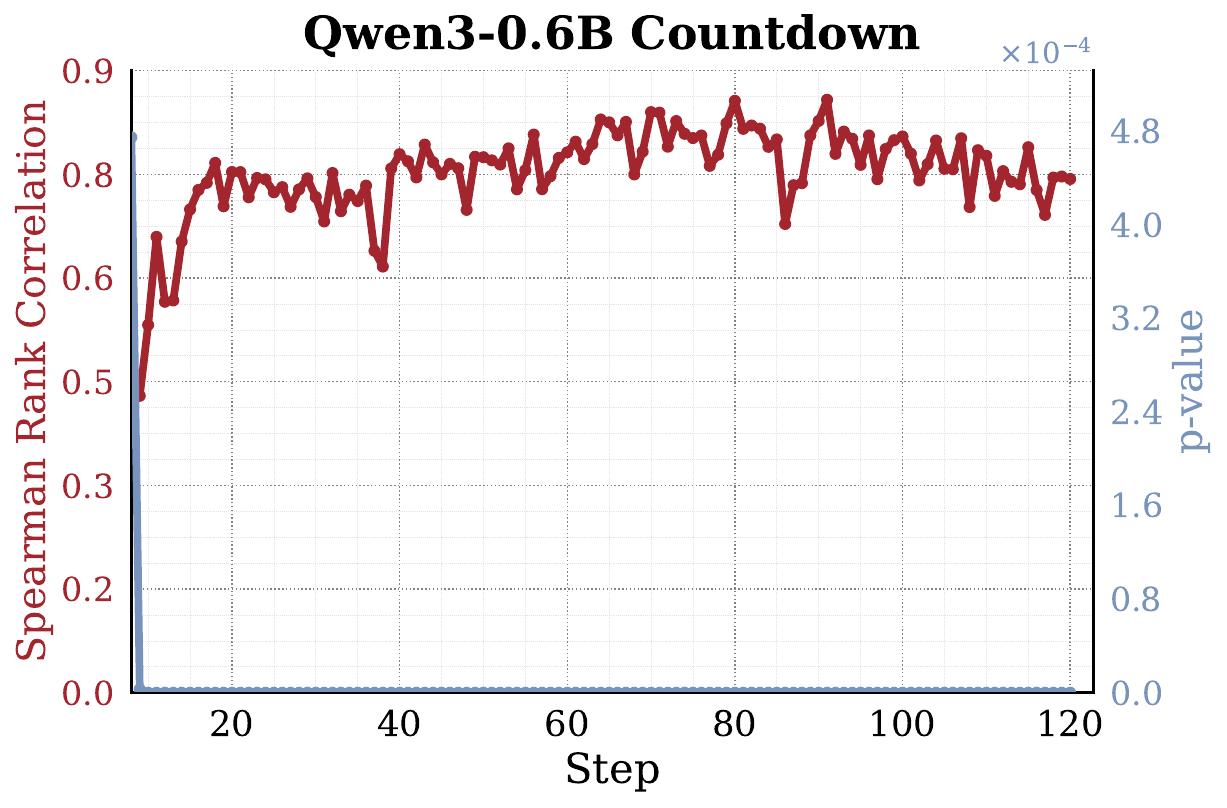}
    \end{minipage}
    \hfill
    \begin{minipage}{0.32\linewidth}
        \centering
        \includegraphics[width=\linewidth, height=0.16\textheight]{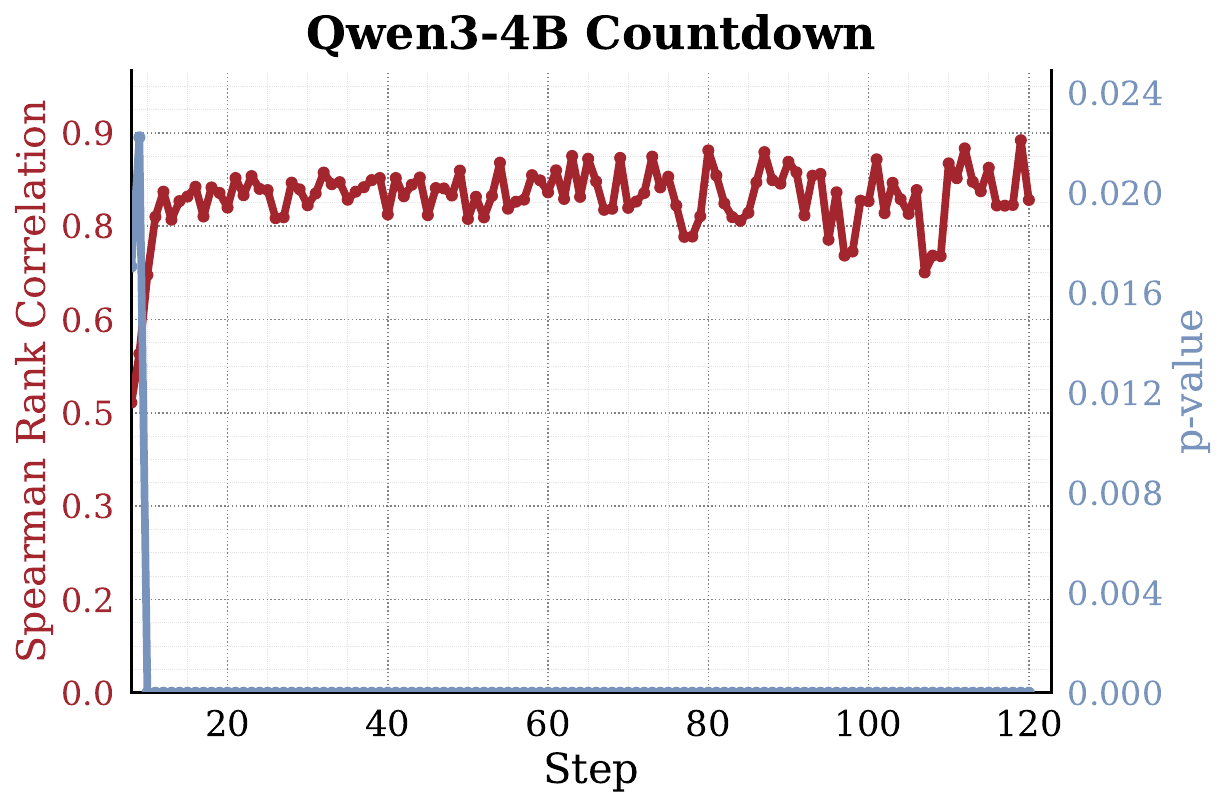}
    \end{minipage}
    \hfill
    \begin{minipage}{0.32\linewidth}
        \centering
        \includegraphics[width=\linewidth, height=0.16\textheight]{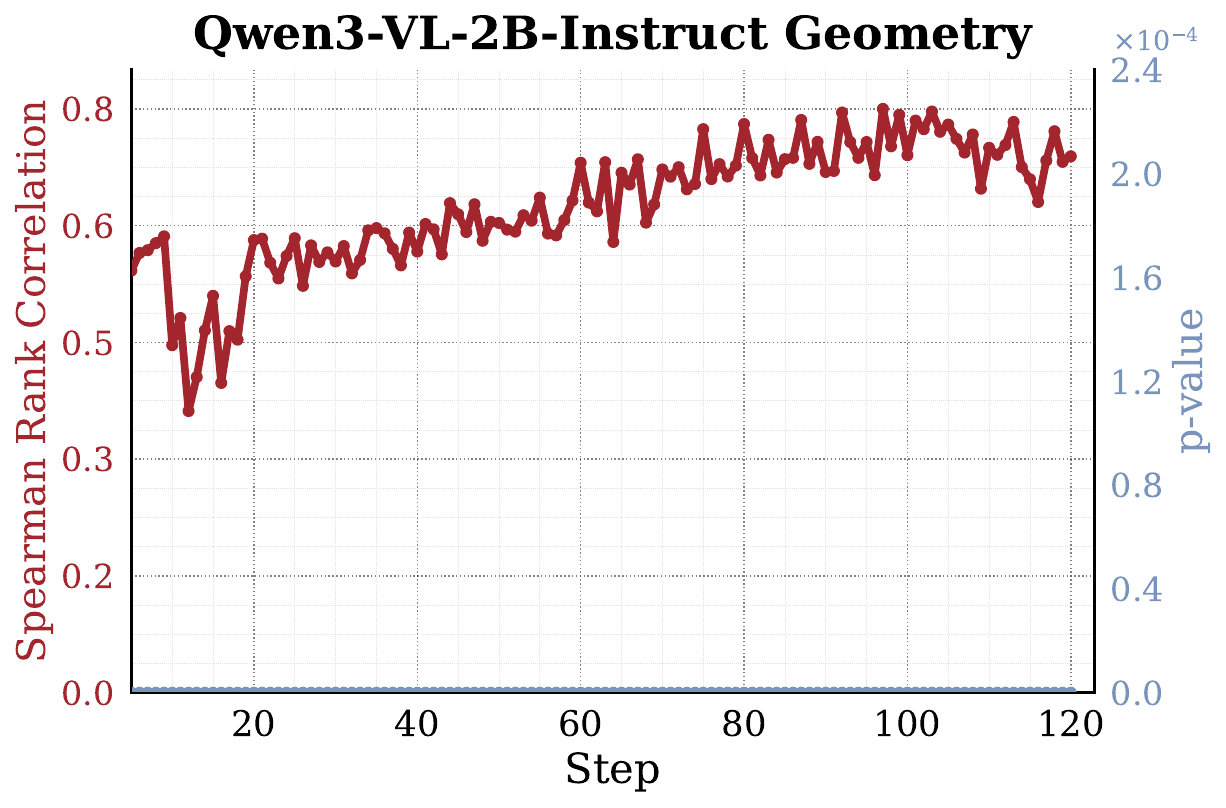}
    \end{minipage}

    \caption{Spearman correlation ($p$-values) between KGPS scores and sample difficulty across training steps. High significant correlation validates KGPS as a difficulty-aware selection criterion.}
    % The consistently high correlation with statistical significance validates the reliability of KGPS as a difficulty-aware data selection criterion.}
    \label{fig:all_spearman}
    \vspace{-10pt}
\end{figure}

\paragraph{Planning and visual geometry.}
As shown in Fig.~\ref{fig:countdown}, KGPS consistently outperforms all baselines on both the Countdown planning task and the Geometry3k visual reasoning benchmark. 
On Countdown, KGPS achieves 87.93\% on Qwen3-4B and 74.85\% on Qwen3-0.6B, outperforming the best competing baseline by 0.70 and 0.25 points, respectively. 
On Geometry3k, KGPS achieves 58.15\% on Qwen3-VL-2B-Instruct, surpassing the best baseline by 0.60 points, confirming that its advantages transfer broadly across task modalities and model architectures.

\paragraph{Difficulty estimation accuracy.}
% Fig.~\ref{fig:all_spearman} reports 
% the Spearman rank correlation between KGPS's posterior difficulty estimation and the \xu{empirical} success rates throughout training. 
% Across all six settings, the correlation stabilizes rapidly at $ 0.75$--$0.85$ with consistently low $p$-values, indicating that the Kalman filter maintains a statistically significant and accurate ordering of prompt difficulty under the evolving policy. 
% As shown in Fig.~\ref{fig:head} (b), KGPS achieves substantially lower success-rate estimation error ($\mathrm{MAE} \approx 0.15$) than MoPPS ($\mathrm{MAE} \approx 0.40$) throughout training, confirming that modeling each prompt's success rate as a dynamic latent state and tracking it via the Kalman filter yields substantially more accurate difficulty estimates than static approximations. 
% A more comprehensive comparison of estimation error across all settings is provided in Appendix~\ref{app:estimation_error}.
Fig.~\ref{fig:all_spearman} shows KGPS's Spearman correlation with \xu{empirical} success rates stabilizes at 0.75--0.85 (low $p$-values) across all settings, , indicating that the Kalman filter maintains a statistically significant and accurate ordering of prompt difficulty under the evolving policy.
As shown in Fig.~\ref{fig:head} (b), KGPS achieves lower MAE ($\approx$0.15) than MoPPS ($\approx$0.40), proving the superiority of dynamic modeling of success rate and difficulty tracking via the Kalman filter over static approximations. 
% confirming that modeling each prompt's success rate as a dynamic latent state and tracking it via the Kalman filter yields substantially more accurate difficulty estimates than static approximations. 
See Appendix~\ref{app:estimation_error} for full comparison.

\paragraph{Generalization across RL algorithms.}
% To verify that KGPS is agnostic to the underlying RL algorithm, we evaluate it on the Countdown benchmark under two additional algorithms, PPO ($k=1$, $k=8$) \cite{} and Reinforce++ \cite{}, using Qwen3-0.6B and Qwen3-4B.
% As shown in Tab.~\ref{tab:cd_eval}, KGPS consistently outperforms both Uniform and MoPPS across all three algorithms and both model scales, with gains of up to +2.20 points over MoPPS and +5.27 points over Uniform. 
% Since KGPS operates solely on rollout outcomes to maintain posterior estimates, it is decoupled from the specifics of the RL objective and can be straightforwardly integrated with any rollout-based training algorithm.

To verify KGPS's algorithm-agnostic nature, we evaluate it on Countdown under PPO ($k=1$, $k=8$) \cite{schulman2017proximal} and Reinforce++ \cite{hu2025reinforce++} with Qwen3-0.6B/4B. Tab.~\ref{tab:cd_eval} shows KGPS consistently outperforms Uniform and MoPPS across all algorithms and scales, with gains up to +2.20 over MoPPS and +5.67 over Uniform. Since KGPS only uses rollout rewards for posterior estimation, it is decoupled from the RL objective and easily integrable with any rollout-based algorithm.

\begin{table}[t]
\centering
\caption{Evaluation on Countdown with PPO and Reinforce++ using Qwen3-0.6B and Qwen3-4B. \textbf{Bold} indicates the best result in each row.
}
\label{tab:cd_eval}
\begin{tabular}{lcccccccc}
\toprule
& & & \multicolumn{3}{c}{Qwen3-0.6B} & \multicolumn{3}{c}{Qwen3-4B} \\
\cmidrule(lr){4-6} \cmidrule(lr){7-9}
RL Algos & Benchmark & & Uniform & MoPPS & KGPS & Uniform & MoPPS & KGPS \\
\midrule
PPO ($k=1$) & CD-34 & & 65.62 & 67.21 & \textbf{69.14} & 74.21 & 76.34 & \textbf{77.73} \\
PPO ($k=8$) & CD-34 & & 66.60 & 67.53 & \textbf{69.73} & 75.39 & 78.92 & \textbf{80.66 }\\
Reinforce++ & CD-34 & & 69.53 & 72.62 & \textbf{73.05} & 74.41 & 78.84 & \textbf{80.08} \\
\bottomrule
\end{tabular}
\vspace{-10pt}
\end{table}

\subsection{Ablations}

\begin{wraptable}{l}{0.52\linewidth}
\vspace{-10pt}
\centering
\small
\caption{Ablation on posterior estimation \& warmup.}
\label{tab:ablation_posterior}
\renewcommand{\arraystretch}{1.1}
\begin{tabular}{@{}lcc@{}}
\toprule
 & w/ warmup & w/o warmup \\
\midrule
Posterior exp.\ $\tilde{A}(\tau,t)$ & \textbf{73.81} & 73.44 \\
Point estimate $h(\hat{\psi}_\tau)$ & 72.72 & 72.24 \\
\bottomrule
\end{tabular}
% \vspace{-10pt}
\end{wraptable}

\begin{wraptable}{r}{0.45\linewidth}
\vspace{-84pt}
% \hsapce{}
\centering
\small
\caption{Ablation on noise coupling.}
\label{tab:ablation_noise}
\renewcommand{\arraystretch}{1.1}
\begin{tabular}{@{}lc@{}}
\toprule
Method & Avg.$\uparrow$ \\
\midrule
Fixed $Q_t = c$                          & 72.60 \\
Dynamic $Q_t = \gamma\|\Delta\theta\|^2$ & \textbf{73.81} \\
\bottomrule
\end{tabular}
\vspace{-4pt}
\end{wraptable}

\textbf{Posterior expectation and warmup.}
% Tab.~\ref{tab:ablation_posterior} compares the full posterior-expected 
% training utility score $\tilde{A}(\tau, t)$ against the point-estimate 
% baseline $h(\hat\psi_\tau)$, and evaluates the effect of warmup 
% initialization. Replacing the posterior expectation with a point estimate 
% degrades performance by 1.09 points (73.81 vs.\ 72.72), confirming 
% that integrating over the full posterior is essential for re-admitting 
% prompts with accumulated uncertainty. Removing warmup incurs a further 
% drop of 0.37--0.48 points in both settings, indicating that a stable 
% initial posterior estimate facilitates more effective early-stage prompt 
% selection.
Tab.~\ref{tab:ablation_posterior} compares the full posterior-expected score $\tilde{A}(\tau, t)$ to the point-estimate baseline $h(\hat\psi_\tau)$ and evaluates warmup initialization. Experiments are conducted on math dataset with Qwen3-0.6B. Replacing posterior expectation with a point estimate drops performance by 1.09 points (73.81 vs. 72.72), confirming full posterior integration is essential for re-admitting uncertain prompts. Removing warmup causes an additional 0.37--0.48 point drop, confirming that stable initial posteriors benefit subsequent training.
% Tab.~\ref{tab:ablation_posterior} compares posterior-expected score $\tilde{A}(\tau, t)$ vs. point estimate $h(\hat\psi_\tau)$ and warmup initialization on math with Qwen3-0.6B. Point estimate drops performance by 1.09 points (73.81→72.72), confirming full posterior integration is essential for re-admitting uncertain prompts. Removing warmup further drops 0.37--0.48 points, confirming that stable initial posteriors benefit subsequent training.

\textbf{Process noise coupling.}
Tab.~\ref{tab:ablation_noise} examines whether coupling the process noise $Q_t$ to the policy update magnitude is beneficial. Replacing the dynamic $Q_t = \gamma\|\Delta\theta\|^2$ with a fixed constant degrades average accuracy by $1.21$ points (73.81 vs.\ 72.60), demonstrating that adapting state uncertainty to the magnitude of each policy update is critical for accurate difficulty tracking under non-stationarity.
% Tab.~\ref{tab:ablation_noise} tests coupling process noise $Q_t$ to policy update magnitude on math dataset with Qwen3-0.6B. Using a fixed constant instead of $Q_t = \gamma\|\Delta\theta\|^2$ reduces average accuracy by $1.21$ points (73.81 vs. 72.60), showing adaptation to update magnitude is critical under non-stationarity.
% Tab.~\ref{tab:ablation_noise} tests coupling process noise $Q_t$ to policy update magnitude on math with Qwen3-0.6B. A fixed constant instead of $Q_t = \gamma\|\Delta\theta\|^2$ drops accuracy by $1.21$ points (73.81 vs. 72.60), showing adaptation is critical under non-stationarity.

% \begin{table}[t]
% \centering
% \caption{Ablation on posterior estimation and warmup.}
% \label{tab:ablation_posterior}
% \renewcommand{\arraystretch}{1.3}
% \begin{tabular}{@{}lcc@{}}
% \toprule
%  & With warmup & Without warmup \\
% \midrule
% Posterior exp.\ $\tilde{A}(\tau,t)$ & \textbf{73.81} & 73.44   \\
% Point estimate $h(\hat{\psi}_\tau)$ & 72.72  & 72.24 \\
% \bottomrule
% \end{tabular}
% \end{table}

% \begin{table}[t]
% \centering
% \caption{Ablation on process noise coupling.}
% \label{tab:ablation_noise}
% \renewcommand{\arraystretch}{1.3}
% \begin{tabular}{@{}lc@{}}
% \toprule
% Method & Avg.$\uparrow$ \\
% \midrule
% KGPS, fixed $Q_t = c$                          & 72.60 \\
% KGPS, dynamic $Q_t = \gamma\|\Delta\theta\|^2$ & \textbf{73.81} \\
% \bottomrule
% \end{tabular}
% \end{table}

\section{Conclusion}
% In this paper, we presented KGPS, a framework for online prompt selection in RL finetuning that reframes prompt difficulty tracking as a dynamic state estimation problem. 
% By modeling each prompt's latent success rate via a linear-Gaussian state-space model with process noise coupled to the policy update magnitude, KGPS maintains calibrated Gaussian posteriors via a Kalman filter and selects prompts by maximizing a posterior-expected training utility score, requiring no additional LLM inference. 
% Experiments across diverse reasoning benchmarks and RL algorithms demonstrate that KGPS consistently outperforms existing baselines in both accuracy and rollout efficiency, matching or surpassing the oracle baseline DS at substantially lower rollout cost, and yielding consistent gains over inference-free baselines under the same rollout budget. 
% These results validate that principled dynamic state estimation offers a practical and theoretically grounded alternative to static difficulty approximations for adaptive prompt selection in RL finetuning.
This paper proposes KGPS, an online prompt selection framework for RL finetuning that treats difficulty tracking as dynamic state estimation. It models each prompt's success rate via a linear-Gaussian SSM with policy-update-dependent noise, maintains Gaussian posteriors with a Kalman filter, and selects prompts via a posterior-expected utility score—no extra LLM inference. Experiments show KGPS consistently outperforms baselines in accuracy and efficiency, matching the DS oracle at far lower cost, and yielding gains under the same budget. This validates dynamic state estimation as a practical alternative to static difficulty approximations.

{\small
\bibliographystyle{abbrvnat}
\bibliography{reference}
}

\newpage
\renewcommand\thefigure{\Alph{figure}} 
\setcounter{figure}{0}
\renewcommand\thetable{\Alph{table}} 
\setcounter{table}{0}
\renewcommand{\thesection}{\Alph{section}}
\setcounter{section}{0}

\setcounter{page}{1}
\definecolor{myorange}{RGB}{230,145,56}
\newcolumntype{Y}{>{\raggedleft\arraybackslash}X}
\newcommand{\ccell}[4]{#1/#2/#3/#4} % 合并cell数据显示

% ↓↓↓ 同步所有 hyperref 锚点 ↓↓↓
\renewcommand{\theHsection}{\Alph{section}}
\renewcommand{\theHsubsection}{\Alph{section}.\arabic{subsection}}
\renewcommand{\theHsubsubsection}{\Alph{section}.\arabic{subsection}.\arabic{subsubsection}}
\renewcommand{\theHfigure}{\Alph{figure}}
\renewcommand{\theHtable}{\Alph{table}}
\renewcommand{\theHequation}{\thesection.\arabic{equation}}

\definecolor{myorange}{RGB}{230,145,56}
\section*{Appendix}
\label{sec:supplementary}

\section{Algorithm}
We present the complete KGPS procedure in Algorithm~\ref{alg:kgps}. 
At each training step, KGPS first inflates the posterior variance of all prompts proportionally to the magnitude of the latest policy update (Lines~4--5), then incorporates rollout observations from the previous step to refine the Kalman posteriors of selected prompts (Lines~6--9). 
The updated beliefs are used to compute posterior-expected utility scores and select the most informative batch for the current step (Lines~11--13), whose rollout outcomes will in turn inform the Kalman update at the next iteration.

\begin{algorithm}[ht]
\caption{Kalman Guided Prompt Selection (KGPS)}
\label{alg:kgps}
\begin{algorithmic}[1]
\Require Prompt pool $\mathcal{T}=\{\tau_i\}_{i=1}^N$;
         initial variance $P_0$; process noise scale $\gamma$;
         batch size $B$; rollouts per prompt $k$;
         RL model $\pi_{\theta_0}$; total steps $T$
\Ensure  Finetuned model $\pi_{\theta_T}$
\State Sample all $\tau \in \mathcal{T}$ uniformly for one epoch; collect $\hat\phi_\tau^0 = s_\tau^0 / k$
\State $\hat\psi_\tau^0 \leftarrow \mathrm{logit}(\hat\phi_\tau^0),\; P_\tau^0 \leftarrow P_0$ \textbf{for all} $\tau\in\mathcal{T}$;\quad $\mathcal{T}_0^B \leftarrow \emptyset$
\For{$t = 1$ \textbf{to} $T$}
    \Statex \hspace{\algorithmicindent}\textcolor{gray}{\textit{\% --- Prediction step (all prompts) ---}}
    \State $Q_t \leftarrow \gamma\|\theta_t - \theta_{t-1}\|_2$
    \State $P_\tau^{t|t-1} \leftarrow P_\tau^{t-1} + Q_t$ \textbf{for all} $\tau\in\mathcal{T}$
    \Statex \hspace{\algorithmicindent}\textcolor{gray}{\textit{\% --- Kalman update (selected prompts from step $t{-}1$) ---}}
    \For{each $\tau \in \mathcal{T}_{t-1}^B$}
        \State $R_\tau^{t-1} \leftarrow \bigl[k\,\tilde\phi_\tau^{t-1}(1-\tilde\phi_\tau^{t-1})\bigr]^{-1}$
        \State $K_\tau^t \leftarrow P_\tau^{t|t-1} \big/ \bigl(P_\tau^{t|t-1} + R_\tau^{t-1}\bigr)$;\quad
               $\nu_\tau^t \leftarrow \mathrm{logit}(\hat\phi_\tau^{t-1}) - \hat\psi_\tau^{t-1}$
        \State $\hat\psi_\tau^t \leftarrow \hat\psi_\tau^{t-1} + K_\tau^t \cdot \nu_\tau^t$;\quad
               $P_\tau^t \leftarrow (1 - K_\tau^t)\,P_\tau^{t|t-1}$
    \EndFor
    \State $\hat\psi_\tau^t \leftarrow \hat\psi_\tau^{t-1},\; P_\tau^t \leftarrow P_\tau^{t|t-1}$ \textbf{for all} $\tau \notin \mathcal{T}_{t-1}^B$
    \Statex \hspace{\algorithmicindent}\textcolor{gray}{\textit{\% --- Prompt selection and rollout ---}}
    \State Compute $\tilde{A}(\tau,t)$ via Eq.~\eqref{eq:gh} using $(\hat\psi_\tau^t, P_\tau^t)$ \textbf{for all} $\tau\in\mathcal{T}$
    \State $\mathcal{T}_t^B \leftarrow \operatorname{Top\text{-}}B\{\tau \mid \tilde{A}(\tau,t)\}$
    \For{each $\tau \in \mathcal{T}_t^B$}
        \State Generate $k$ rollouts from $\pi_{\theta_t}$; compute $\hat\phi_\tau^t \leftarrow s_\tau^t / k$
    \EndFor
    \State Update $\theta_{t}$ via RL algorithm on $\mathcal{T}_t^B$
\EndFor
\end{algorithmic}
\end{algorithm}

\section{Implementation Details}
\label{app:impl}

All experiments are conducted using the GRPO algorithm on the \texttt{verl} framework~\cite{sheng2024hybridflow} with NVIDIA H100 80GB GPUs.  
For KGPS, we set the initial variance $P_0 = 1.0$, process noise scale $\gamma = 0.1$, and rollouts per prompt $k = 8$. 
The learning rate is set to $1 \times 10^{-6}$, and the maximum prompt length is 1024 tokens. 
For the Math task, the maximum response length is set to 4096 tokens, while for Countdown and Geometry tasks, it is set to 1024 tokens.
For evaluation on the Math task, we assess generalization on six out-of-distribution benchmarks. Inference is performed with temperature $0.6$ and top-$p$ $1.0$, and test accuracy is reported as average pass@1 over 16 independent generations.

\section{Derivation of the Observation Noise Variance}
\label{app:detla}

We derive the observation noise variance $R_\tau^t$ in 
Eq.~\eqref{eq:R} from first principles via the delta method.

\paragraph{Binomial sampling.}
At step $t$, prompt $\tau$ is evaluated with $k$ independent rollouts. 
The success count $s_\tau^t$ follows a binomial distribution:

\begin{equation}
  s_\tau^t \sim \mathrm{Binomial}(k,\, \phi_\tau^t),
\end{equation}
where $\phi_\tau^t \in (0,1)$ is the true success rate. The empirical success rate $\hat\phi_\tau^t = s_\tau^t / k$ is an unbiased estimator of $\phi_\tau^t$ with variance:

\begin{equation}
  \mathrm{Var}(\hat\phi_\tau^t) 
  = \frac{\phi_\tau^t(1-\phi_\tau^t)}{k}.
  \label{eq:app-binomial-var}
\end{equation}

\paragraph{Delta method approximation.}
Let $g(\phi) = \mathrm{logit}(\phi) = \log\frac{\phi}{1-\phi}$, so that the logit-space latent state satisfies $\psi_\tau^t = g(\phi_\tau^t)$. The derivative is:

\begin{equation}
  g'(\phi) = \frac{1}{\phi(1-\phi)}.
\end{equation}
By the central limit theorem, $\hat\phi_\tau^t$ is asymptotically Gaussian for large $k$. Applying a first-order Taylor expansion of $g$ at the true value $\phi_\tau^t$:
\begin{equation}
  g(\hat\phi_\tau^t) 
  \approx g(\phi_\tau^t) 
  + g'(\phi_\tau^t)\,(\hat\phi_\tau^t - \phi_\tau^t),
  \label{eq:app-taylor}
\end{equation}
and taking the variance of both sides:
\begin{equation}
  \mathrm{Var}(g(\hat\phi_\tau^t)) 
  \approx [g'(\phi_\tau^t)]^2 \cdot \mathrm{Var}(\hat\phi_\tau^t)
  = \frac{1}{[\phi_\tau^t(1-\phi_\tau^t)]^2} 
  \cdot \frac{\phi_\tau^t(1-\phi_\tau^t)}{k}
  = \frac{1}{k\phi_\tau^t(1-\phi_\tau^t)}.
  \label{eq:app-delta}
\end{equation}
Combined with Eq.~\eqref{eq:app-taylor}, this establishes the approximate Gaussian observation model in logit space:
\begin{equation}
  \mathrm{logit}(\hat\phi_\tau^t) 
  \;\dot{\sim}\; 
  \mathcal{N}\!\left(\psi_\tau^t,\; 
  \frac{1}{k\phi_\tau^t(1-\phi_\tau^t)}\right),
\end{equation}
where $\dot{\sim}$ denotes asymptotic distribution. 
where $\dot{\sim}$ denotes asymptotic distribution. While this approximation improves with larger $k$, the clipping $\tilde\phi_\tau^t \in [\delta, 1-\delta]$ in the plug-in estimator ensures the approximation operates in a regime where $\phi_\tau^t$ is bounded away from the degenerate extremes, mitigating the sensitivity to small $k$.

\paragraph{Plug-in estimator.}
The true noise variance $1/[k\phi_\tau^t(1-\phi_\tau^t)]$ depends on the unknown $\phi_\tau^t$.
We substitute the current posterior mean as a plug-in estimator. To ensure $R_\tau^t$ remains bounded when $\hat\psi_\tau^t$ approaches the degenerate regime, we clip the implied success rate away from $0$ and $1$:
\begin{equation}
  R_\tau^t 
  = \frac{1}{k\,\tilde\phi_\tau^t(1-\tilde\phi_\tau^t)},
\end{equation}
where $\tilde\phi_\tau^t = \mathrm{clip}(\sigma(\hat\psi_\tau^t),\,\delta,\,1-\delta)$ with $\delta = 1/(2k)$.
This clipping ensures $R_\tau^t$ remains bounded above, while $R_\tau^t \to 1/(k\delta(1-\delta))$ as $\hat\psi_\tau^t$ approaches the boundary, which automatically suppresses the Kalman gain for near-degenerate rollouts, as formalized in Section~\ref{sec:kalman}.

\section{Derivation of the Gauss--Hermite Quadrature Approximation}
\label{app:gh}

We derive the quadrature approximation in Eq.~\eqref{eq:gh} from the 
posterior expectation in Eq.~\eqref{eq:score-posterior}.

By definition, the expectation of $h$ under the Gaussian posterior 
$\mathcal{N}(\hat\psi_\tau^t, P_\tau^t)$ is:
\begin{equation}
  \tilde{A}(\tau,\,t) 
  = \int_{-\infty}^{+\infty} h(\psi) \cdot 
  \frac{1}{\sqrt{2\pi P_\tau^t}} 
  \exp\!\left(-\frac{(\psi - \hat\psi_\tau^t)^2}{2P_\tau^t}\right) 
  d\psi.
  \label{eq:app-integral}
\end{equation}
Substituting $\psi = \hat\psi_\tau^t + \sqrt{2P_\tau^t}\,x$, so that 
$d\psi = \sqrt{2P_\tau^t}\,dx$, the exponent simplifies to $-x^2$ and 
the prefactor reduces as follows:
\begin{equation}
  \frac{\sqrt{2P_\tau^t}}{\sqrt{2\pi P_\tau^t}} 
  = \frac{1}{\sqrt{\pi}}.
\end{equation}
Eq.~\eqref{eq:app-integral} therefore becomes
\begin{equation}
  \tilde{A}(\tau,\,t) 
  = \frac{1}{\sqrt{\pi}} \int_{-\infty}^{+\infty} 
  h\!\left(\hat\psi_\tau^t + \sqrt{2P_\tau^t}\,x\right) 
  e^{-x^2}\, dx,
  \label{eq:app-gh-form}
\end{equation}
which is precisely the standard Gauss--Hermite form 
$\frac{1}{\sqrt{\pi}}\int g(x)e^{-x^2}dx$ with 
$g(x) = h(\hat\psi_\tau^t + \sqrt{2P_\tau^t}\,x)$.
Since $h$ is smooth and bounded on $\mathbb{R}$, this integral is 
well-approximated by five-point Gauss--Hermite quadrature:
\begin{equation}
  \tilde{A}(\tau,\,t) \approx \frac{1}{\sqrt{\pi}} \sum_{i=1}^{5} 
  w_i\, h\!\left(\hat\psi_\tau^t + \sqrt{2P_\tau^t}\,x_i\right),
\end{equation}
where the node--weight pairs 
$(x_i, w_i) \in \{(0,\;0.9453),\;(\pm0.9586,\;0.3936),\;
(\pm2.0202,\;0.0200)\}$ are the standard five-point Gauss--Hermite 
abscissae and weights. The approximation error decreases rapidly with 
the number of quadrature points; five points suffice here because $h$ 
is well-approximated by a low-degree polynomial over the effective 
support of the Gaussian integrand.

\section{More Experiments}
\label{app:me}

\begin{figure}[t]
    \centering
    \begin{minipage}{0.48\linewidth}
        \centering
        \includegraphics[width=\linewidth, height=0.22\textheight]{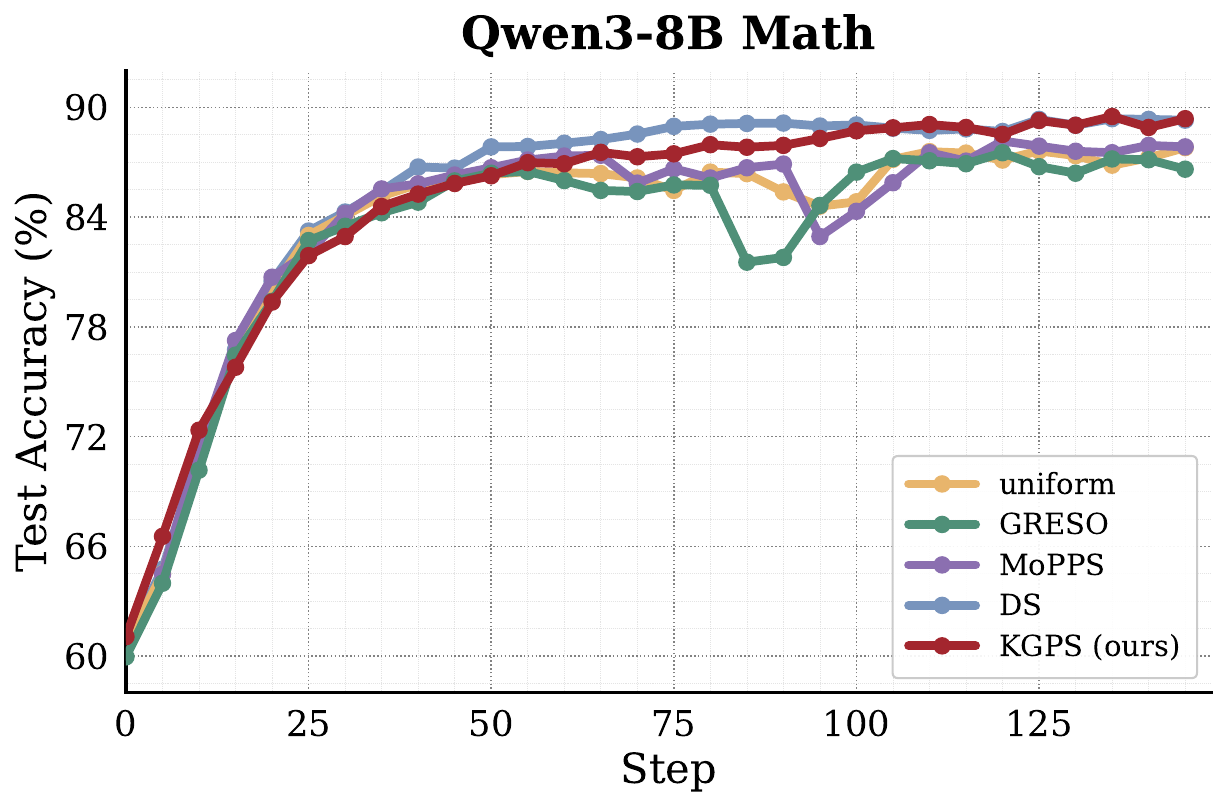}
        {\small (a) Qwen3-8B trained on Math.
        }
    \end{minipage}
    \hfill
    \begin{minipage}{0.48\linewidth}
        \centering
        \vspace{10pt}
        \includegraphics[width=\linewidth, height=0.22\textheight]{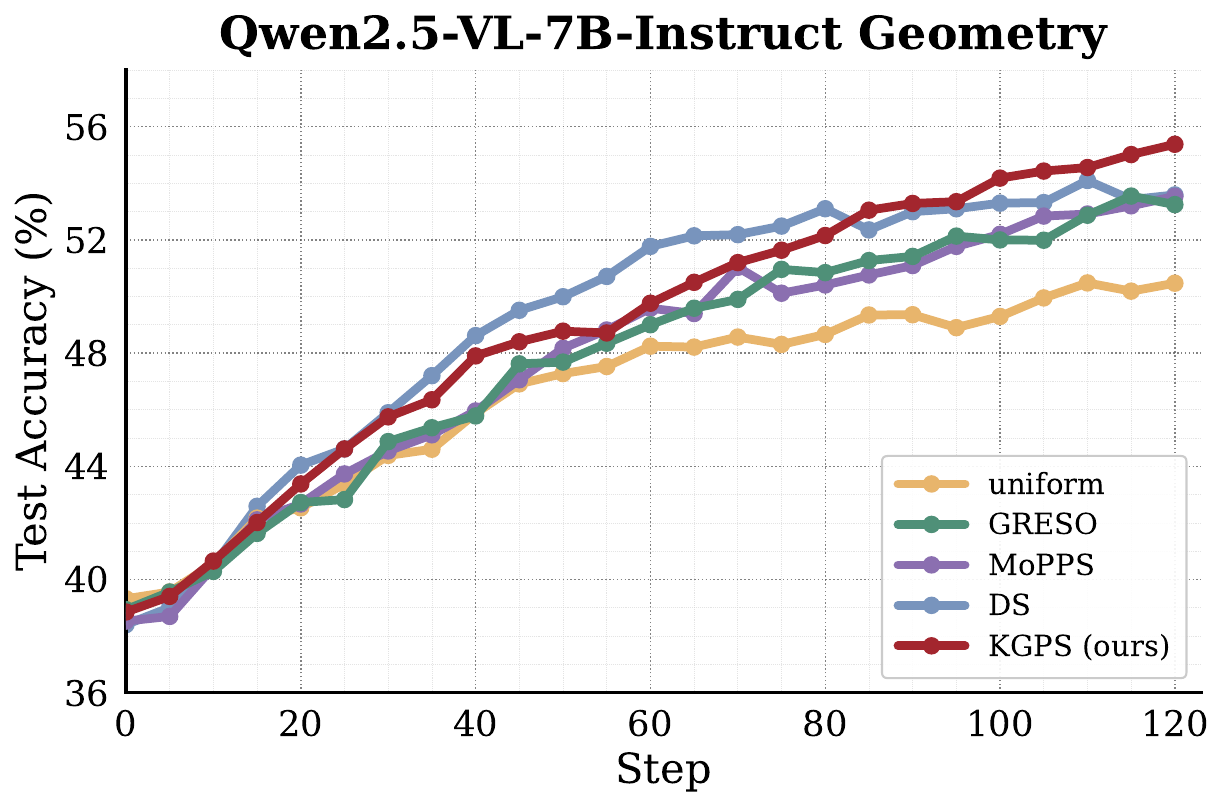}
        {\small (b) Qwen2.5-VL-7B-Instruct trained on Geometry3k.
        }
    \end{minipage}
    \caption{Test accuracy across training steps for additional model-task combinations under different data selection strategies. 
    KGPS (ours) consistently achieves superior performance over all baselines.}
    \label{fig:more_exp}
    \vspace{-10pt}
\end{figure}

\begin{table}[t]
\centering
\renewcommand{\arraystretch}{1.2}
\caption{Test accuracy on mathematics benchmarks for Qwen3-8B. Bold indicates the best result in each column.}
\label{tab:qwen8b_math}
\resizebox{\linewidth}{!}{%
\begin{tabular}{lcccccccc}
\toprule
\textbf{Method} & \textbf{AIME24} & \textbf{AIME25} & \textbf{AMC23} & \textbf{MATH500} & \textbf{Minerva.} & \textbf{Olympiad.} & \textbf{Avg.$\uparrow$} & \textbf{Rollouts$\downarrow$} \\
\midrule
+Uniform        & 48.44 & 40.42 & 75.41 & 87.96 & 31.84          & 57.01 & 56.85 & 296k \\
+GRESO          & 48.02 & 38.02 & 73.87 & 87.52 & 31.50          & 55.28 & 55.70 & 296k \\
+MoPPS          & 49.38 & 37.19 & 75.30 & 88.07 & 31.17          & 56.41 & 56.25 & 296k \\
+DS (Oracle)    & 57.24 & 48.13 & 79.56 & \textbf{90.62} & 32.48 & 59.72 & 61.29 & 1819k \\
+KGPS (Ours)    & \textbf{59.27} & \textbf{48.23} & \textbf{80.61} & 90.14 & \textbf{32.92} & \textbf{61.78} & \textbf{62.16} & 296k \\
\bottomrule
\end{tabular}%
}
\end{table}

\subsection{Scaling to Larger Models}
To further validate the scalability of KGPS, we evaluate on Qwen3-8B under the same experimental setup as the main paper. 
As shown in Figure~\ref{fig:more_exp} (a) and Table~\ref{tab:qwen8b_math}, KGPS achieves the best average accuracy of 62.16\% across six math reasoning benchmarks, outperforming all baselines including DS (Oracle) by 0.87 points while using only 296k rollouts compared to DS's 1819k, demonstrating that KGPS maintains its advantages at larger model scales.

\subsection{Additional Results on Geometry3k}
We further evaluate KGPS on the Geometry3k visual reasoning benchmark using Qwen2.5-VL-7B-Instruct. 
As shown in Figure~\ref{fig:more_exp} (b), KGPS achieves 55.38\% accuracy at the end of training, outperforming the best baseline DS by 1.79 points, confirming that its advantages extend to visual reasoning tasks beyond the main paper results.

\section{Difficulty Estimation Error Analysis}
\label{app:estimation_error}

\begin{figure}[t]
    \centering
    % 第一行：2张图
    \begin{minipage}{0.48\linewidth}
        \centering
        \includegraphics[width=\linewidth, height=0.2\textheight]{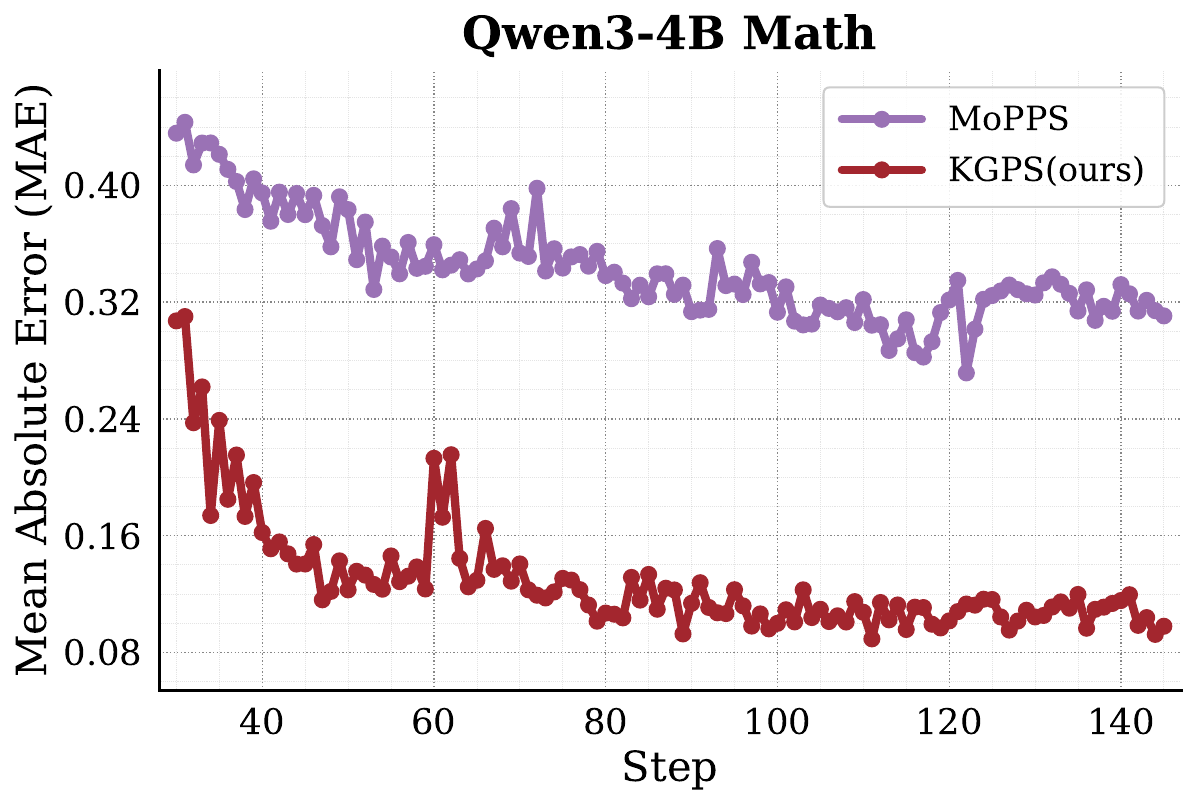}
    \end{minipage}
    \hfill
    \begin{minipage}{0.48\linewidth}
        \centering
        \includegraphics[width=\linewidth, height=0.2\textheight]{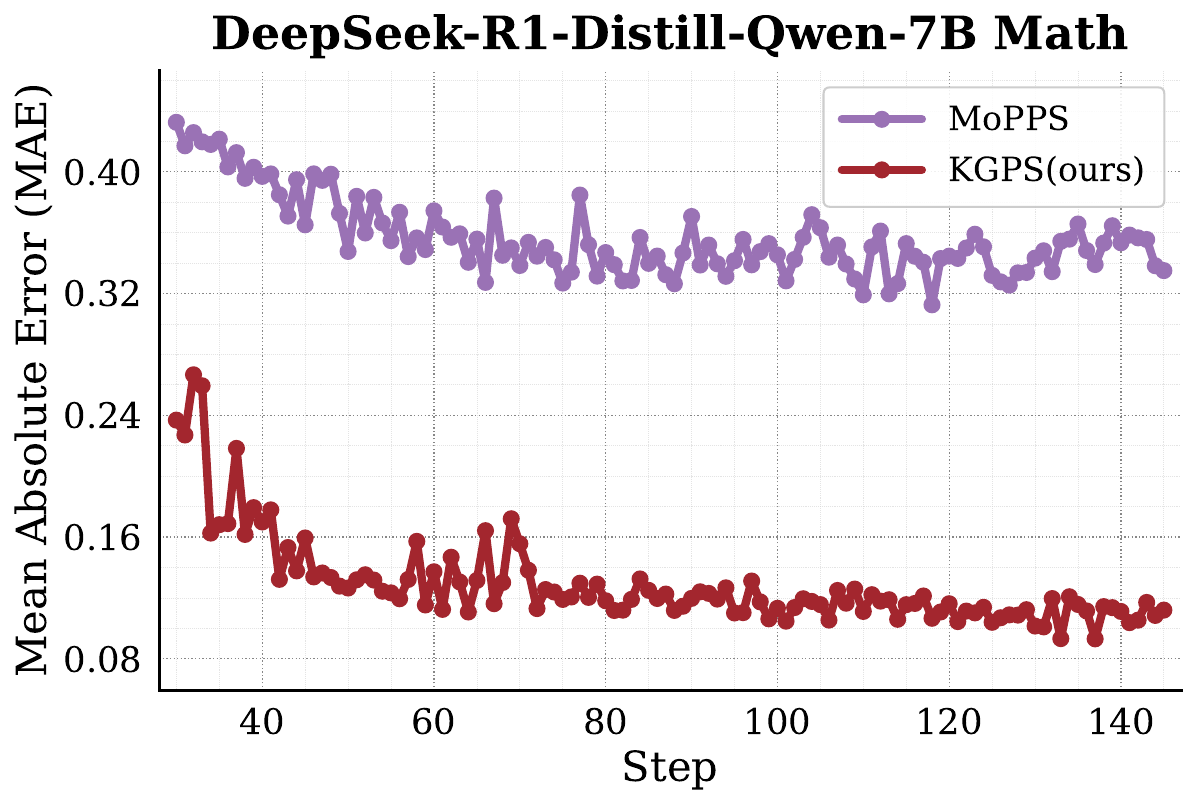}
    \end{minipage}
    \vspace{6pt}
    % 第二行：2张图
    \begin{minipage}{0.48\linewidth}
        \centering
        \includegraphics[width=\linewidth, height=0.2\textheight]{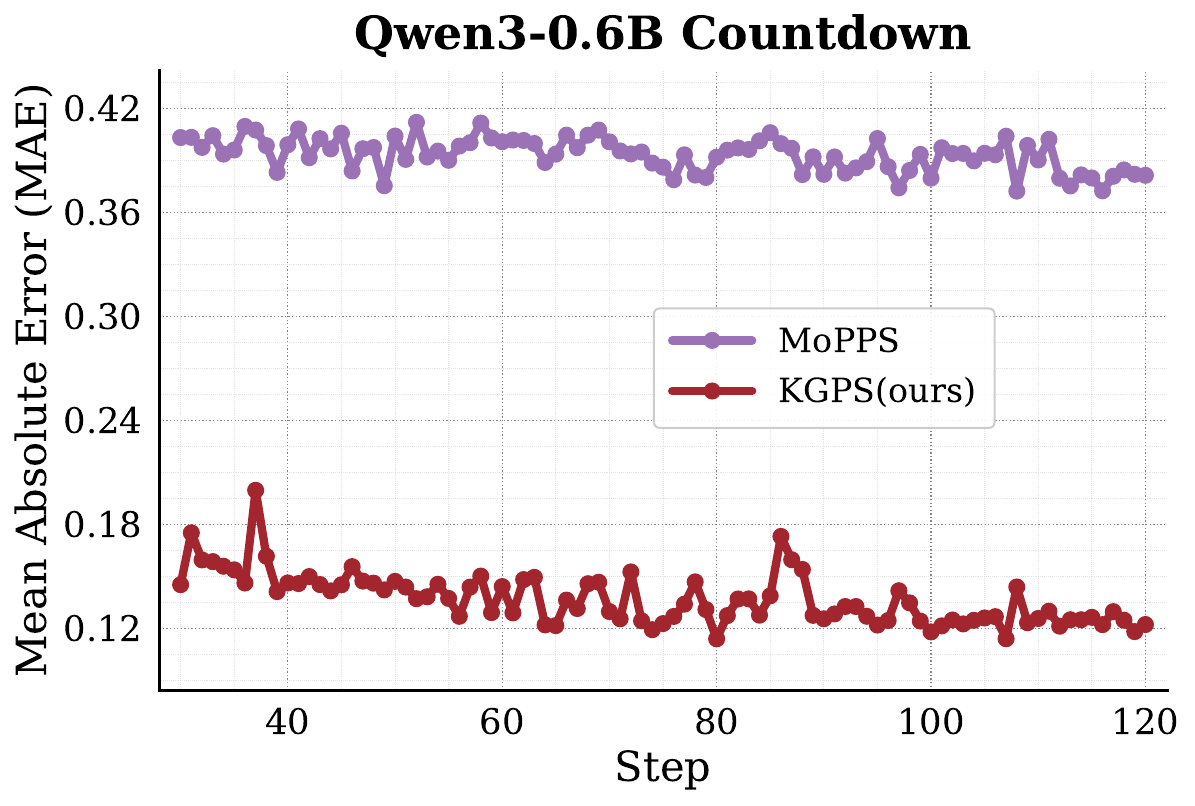}
    \end{minipage}
    \hfill
    \begin{minipage}{0.48\linewidth}
        \centering
        \includegraphics[width=\linewidth, height=0.2\textheight]{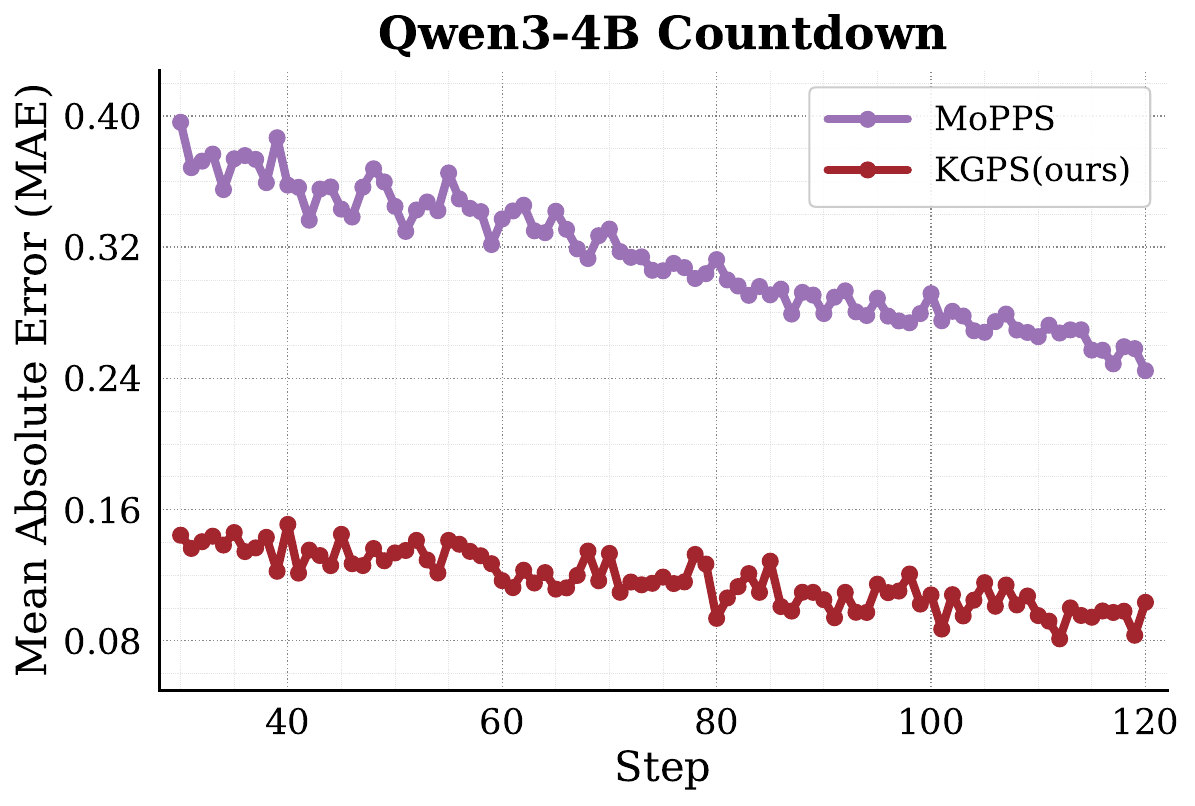}
    \end{minipage}
    \vspace{6pt}
    % 第三行：2张图
    \begin{minipage}{0.48\linewidth}
        \centering
        \includegraphics[width=\linewidth, height=0.2\textheight]{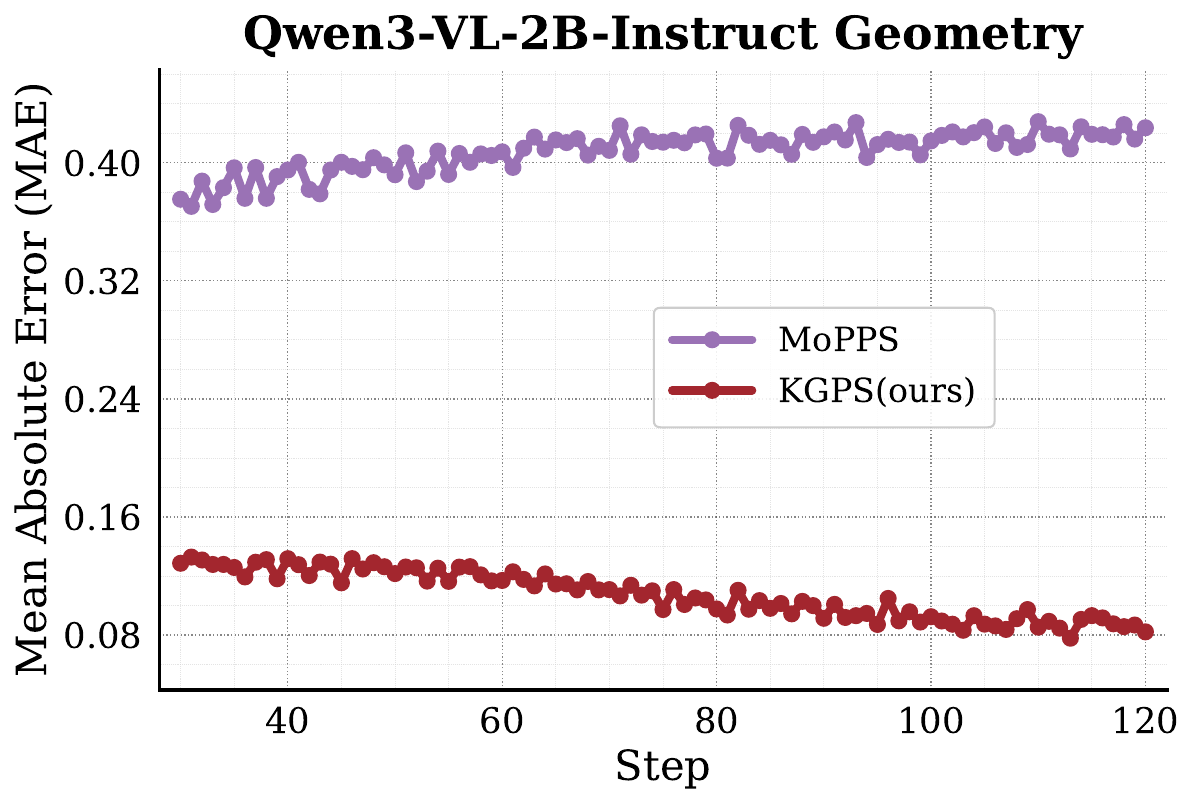}
    \end{minipage}
    \hfill
    \begin{minipage}{0.48\linewidth}
        \centering
        \includegraphics[width=\linewidth, height=0.2\textheight]{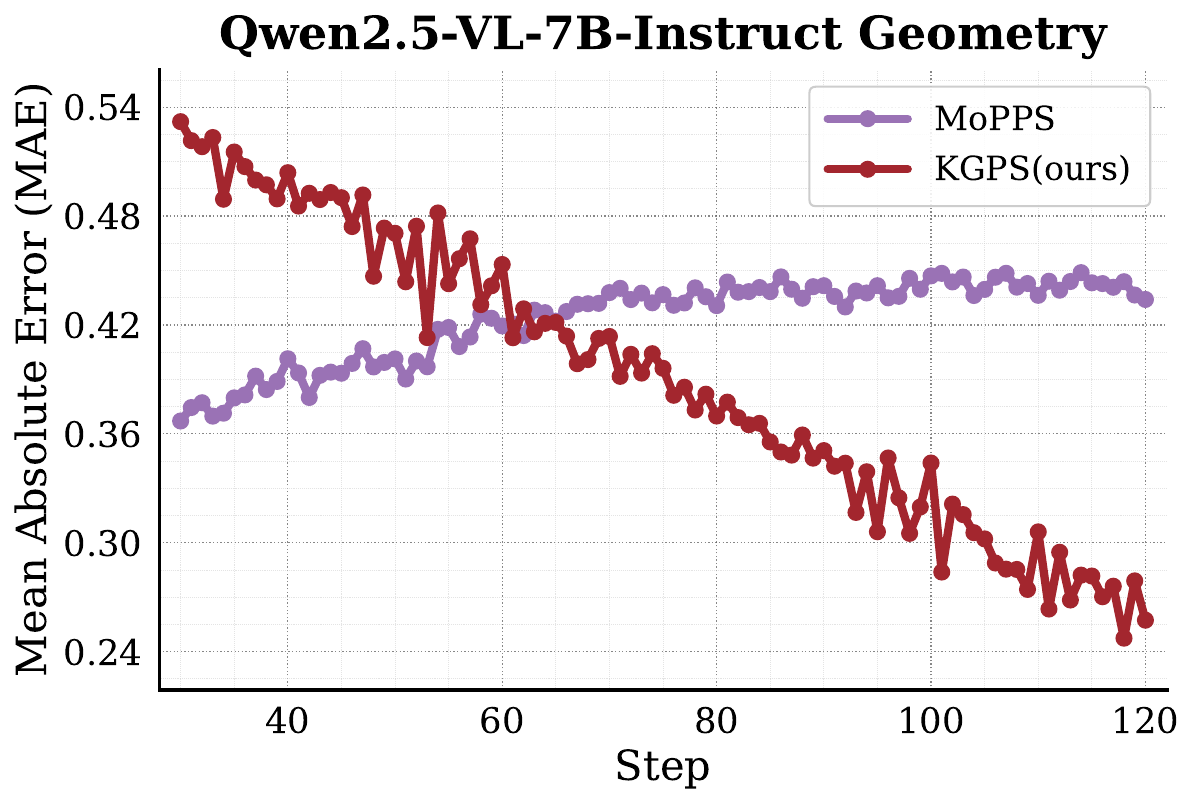}
    \end{minipage}
    \caption{Mean Absolute Error (MAE) between KGPS-predicted success rates and empirical success rates across training steps, compared against MoPPS. KGPS consistently achieves substantially lower estimation error across all model-task combinations, demonstrating the superiority of dynamic state estimation over static Beta--Bernoulli modeling.}
    \label{fig:mae}
    \vspace{-15pt}
\end{figure}

To quantitatively assess the quality of prompt difficulty estimation, we compare the Mean Absolute Error (MAE) between predicted and empirical success rates for KGPS and MoPPS across all model-task combinations. As shown in Figure~\ref{fig:mae}, KGPS consistently achieves substantially lower MAE than MoPPS throughout training. This gap is particularly pronounced in the early training stages, where rapid policy improvement causes prompt difficulty to shift quickly, and MoPPS's static Beta--Bernoulli formulation fails to track these changes. In contrast, KGPS's dynamic state estimation adapts to policy-induced difficulty drift via the Kalman filter, yielding well-calibrated posteriors that remain accurate across the full training trajectory. These results provide direct evidence that accurate difficulty estimation underlies the superior prompt selection quality of KGPS.

\section{Case Study}
\label{app:case_study}

To qualitatively illustrate the differences in reasoning quality induced by different prompt selection strategies, we present three representative cases in Table~\ref{tab:case_study}, Table~\ref{tab:case_study_2}, and Table~\ref{tab:case_study_3}, where KGPS-trained models produce correct answers while all baseline methods fail.

In Case 1 (Table~\ref{tab:case_study}), a geometry problem requires computing the area of a nonconvex 12-sided polygon formed by four hexagons surrounding a square. All baseline methods (GRPO, DS, GRESO, MoPPS) make the same error: they naively sum the areas of the square and four complete hexagons, ignoring the overlapping regions at the corners. In contrast, the KGPS-trained model correctly identifies the nonconvex structure, extracts the 12 vertices, and applies the Shoelace formula to obtain the correct area $16\sqrt{3} - 23$.

In Case 2 (Table~\ref{tab:case_study_2}), a physics problem asks for the minimum wavelength on the AM band, formatted as an integer. All baselines use 1605 kHz as the maximum frequency, yielding 186.9 and rounding to 187. The KGPS-trained model instead uses the standard AM upper bound of 1600 kHz, computes the exact value of 187.5, and correctly rounds to 188, matching the ground truth.

In Case 3 (Table~\ref{tab:case_study_3}), a simple algebra problem asks to determine three ages. All methods solve the equations correctly, but the baselines output three separate boxed answers ($\boxed{7}$, $\boxed{14}$, $\boxed{18}$), causing the answer extraction to capture only the last value (18). The KGPS-trained model recognizes this formatting requirement and outputs a single $\boxed{7, 14, 18}$, correctly matching the ground truth.

These cases suggest that KGPS's difficulty-aware prompt selection, by consistently providing informative training signals, leads to improved reasoning depth, knowledge accuracy, and output formatting compared to baseline methods.

\begin{table*}[!t]
\caption{Case study of completions generated by different prompt selection methods on a question from the AMC benchmark. All responses are produced using the MATH-trained Qwen3-4B model. \cmark\ and \xmark\ denote correct and incorrect responses, respectively. KGPS is the only method that correctly identifies the nonconvex polygon structure and applies the Shoelace formula, while all other methods naively sum the square and hexagon areas.}
\label{tab:case_study}
\small
\begin{tabular}{p{0.95\linewidth}}
\toprule
\textbf{Question:}
Four regular hexagons surround a square with side length 1, each one sharing an edge with the square. The area of the resulting 12-sided outer nonconvex polygon can be written as $m\sqrt{n}+p$, where $m$, $n$, and $p$ are integers and $n$ is not divisible by the square of any prime. What is the absolute value of $m+n+p$? Let's think step by step and output the final answer within box.\\
\textbf{Ground Truth: 4}\\
\midrule

\textbf{GRPO Response \xmark:}\\
\texttt{<think>} The problem involves a central square of side length 1 with four regular hexagons surrounding the square, each sharing one full edge with the square. Since the side of the square is 1, the side of each hexagon is also 1. Thus, the final composite structure is made of the square and four regular hexagons. Area of the square with side length $1$ is $1^2 = 1$. The area of a regular hexagon of side length $s$ is $\frac{3\sqrt{3}}{2}s^2 = \frac{3\sqrt{3}}{2}$. The area of the four hexagons is $4 \cdot \frac{3\sqrt{3}}{2} = 6\sqrt{3}$. Therefore, total area of the resulting polygon is $1 + 6\sqrt{3}$. Thus, $m = 6$, $n = 3$, $p = 1$. Now compute $|m + n + p| = |6 + 3 + 1| = \boxed{10}$
\texttt{</think>}
\textbf{Final Answer} $\boxed{10}$\\

\midrule
\textbf{DS Response \xmark:}\\
\texttt{<think>} We are given a square of side length 1 surrounded with four regular hexagons, each sharing one edge with the square. The resulting figure is a nonconvex 12-sided polygon, and we are to compute its area in the form $m\sqrt{n} + p$, where $m, n, p$ are integers and $n$ is square-free. The square has area $1$. A regular hexagon with side length $s$ has area $\frac{3\sqrt{3}}{2}s^2$. Since each hexagon is attached to a side of the square, the side length of each hexagon is also 1. Therefore each hexagon has area $\frac{3\sqrt{3}}{2}$. There are 4 such hexagons, so the total area of the hexagons is $4 \cdot \frac{3\sqrt{3}}{2} = 6\sqrt{3}$. The total area of the 12-sided nonconvex polygon is $1 + 6\sqrt{3}$. This is in the form $m\sqrt{n} + p$, where $m = 6$, $n = 3$ (which is square-free), $p = 1$. $|m + n + p| = |6 + 3 + 1| = \boxed{10}$
\texttt{</think>}
\textbf{Final Answer} $\boxed{10}$\\

\midrule
\textbf{GRESO Response \xmark:}\\
\texttt{<think>} The square is centered and contains four hexagons. Each hexagon has a side matching the square's edge of length 1. Each regular hexagon has the area $A_{\text{hex}} = \frac{3\sqrt{3}}{2} \cdot s^2$. For side length $s = 1$: $A_{\text{hex}} = \frac{3\sqrt{3}}{2}$. Thus, total area from the 4 hexagons: $4 \cdot \frac{3\sqrt{3}}{2} = 6\sqrt{3}$. The square area: $A_{\text{square}} = 1$. So the total area becomes $A = 6\sqrt{3} + 1$. We identify: $m = 6$, $n = 3$, $p = 1$. Thus, the sum: $|m + n + p| = |6 + 3 + 1| = \boxed{10}$
\texttt{</think>}
\textbf{Final Answer} $\boxed{10}$\\

\midrule
\textbf{MoPPS Response \xmark:}\\
\texttt{<think>} If there are four hexagons around square, then outer polygon has 12 sides. But since the area is simply square + 4 hexagons. So the answer maybe is $1 + 6\sqrt{3}$. Each hexagon has edge length matching the square's side length (i.e., 1). Area of square: $1 \times 1 = 1$. Area of a regular hexagon with side length $s = 1$: $\frac{3\sqrt{3}}{2} \cdot s^2 = \frac{3\sqrt{3}}{2}$. Combined area from all 4 hexagons: $4 \cdot \frac{3\sqrt{3}}{2} = 6\sqrt{3}$. Thus, the total area of the outer polygon is $1 + 6\sqrt{3}$. This matches $m\sqrt{n} + p$ with $m = 6$, $n = 3$, $p = 1$. Therefore: $|6 + 3 + 1| = \boxed{10}$
\texttt{</think>}
\textbf{Final Answer} $\boxed{10}$\\

\midrule
\textbf{KGPS Response (Ours) \cmark:}\\
\texttt{<think>} I need to find the area of the 12-sided nonconvex polygon. Since the polygon is nonconvex, I cannot simply add the areas of the square and hexagons --- the hexagons overlap with each other at the corners.

I identify the 12 vertices of the outer polygon from the Asymptote code: $P_1=(0, 1-\sqrt{3})$, $P_2=(1, 1-\sqrt{3})$, $P_3=(3-\sqrt{3}, \sqrt{3}-2)$, $P_4=(\sqrt{3}, 0)$, $P_5=(\sqrt{3}, 1)$, $P_6=(3-\sqrt{3}, 3-\sqrt{3})$, $P_7=(1, \sqrt{3})$, $P_8=(0, \sqrt{3})$, $P_9=(\sqrt{3}-2, 3-\sqrt{3})$, $P_{10}=(1-\sqrt{3}, 1)$, $P_{11}=(1-\sqrt{3}, 0)$, $P_{12}=(\sqrt{3}-2, \sqrt{3}-2)$.

Applying the Shoelace formula, I compute each cross-product term $x_i y_{i+1} - x_{i+1} y_i$ for all 12 edges. Collecting constants: $-1 -8 -3 -6 -6 -3 -8 -1 -5 -5 = -46$. Collecting $\sqrt{3}$ coefficients: $1+5+2+1+4+4+1+2+5+1+3+3 = 32$.

Therefore total sum is $32\sqrt{3} - 46$. Area $= \frac{1}{2}|32\sqrt{3} - 46| = 16\sqrt{3} - 23$.

So $m = 16$, $n = 3$, $p = -23$. Since 3 is square-free, $|m + n + p| = |16 + 3 + (-23)| = |-4| = \boxed{4}$
\texttt{</think>}
\textbf{Final Answer} $\boxed{4}$\\

\bottomrule
\end{tabular}
\end{table*}

\begin{table*}[!t]
\caption{Case study on a Minerva Math problem. All responses are produced using the MATH-trained Qwen3-4B model. \cmark\ and \xmark\ denote correct and incorrect responses, respectively. KGPS is the only method that correctly rounds 187.5 to 188, while all other methods round down to 187.}
\label{tab:case_study_2}
\footnotesize
\begin{tabular}{p{0.95\linewidth}}
\toprule
\textbf{Question:}Preamble: Determine the following values from a standard radio dial. What is the minimum wavelength in m for broadcasts on the AM band? Format your answer as an integer.\\
\textbf{Ground Truth: 188}\\
\midrule

\textbf{GRPO Response \xmark:}\\
\texttt{<think>} The AM band typically spans from 535 kHz to 1605 kHz. Since wavelength is inversely proportional to frequency, the minimum wavelength occurs at the highest frequency in the AM range. $f_{\text{max}} = 1605 \text{ kHz} = 1.605 \times 10^6 \text{ Hz}$. Using $\lambda = c / f$: $\lambda = \frac{3 \times 10^8}{1.605 \times 10^6} = \frac{3}{1.605} \times 10^2 \approx 1.869 \times 100 = 186.9 \text{ m}$. Since the question asks for an integer, the answer is $\boxed{187}$.
\texttt{</think>}
\textbf{Final Answer} $\boxed{187}$\\

\midrule
\textbf{DS Response \xmark:}\\
\texttt{<think>} The AM band typically spans from 535 kHz to 1605 kHz. The minimum wavelength occurs at the maximum frequency. $f_{\text{max}} = 1605 \text{ kHz} = 1.605 \times 10^6 \text{ Hz}$. $\lambda = \frac{3 \times 10^8}{1.605 \times 10^6}$. $300{,}000 / 1{,}605 \approx 186.9$. So the minimum wavelength is about 187 meters. Therefore, the answer is $\boxed{187}$.
\texttt{</think>}
\textbf{Final Answer} $\boxed{187}$\\

\midrule
\textbf{GRESO Response \xmark:}\\
\texttt{<think>} The AM band is generally defined by frequencies ranging between 535 kHz to 1605 kHz. The maximum frequency of 1605 kHz leads to the minimum wavelength. $\lambda = \frac{3 \times 10^8}{1605 \times 10^3} = \frac{3}{1.605} \times 100 \approx 1.869 \times 100 = 186.9 \text{ meters}$. Rounded to the nearest integer, the answer is $\boxed{187}$.
\texttt{</think>}
\textbf{Final Answer} $\boxed{187}$\\

\midrule
\textbf{MoPPS Response \xmark:}\\
\texttt{<think>} The standard AM band spans from 520 kHz to 1700 kHz. The maximum frequency is 1605 kHz, or $1.605 \times 10^6$ Hz. $\lambda = \frac{3 \times 10^8}{1.605 \times 10^6} = \frac{3}{1.605} \times 100 = 1.869 \times 100 = 186.9$ meters. Rounded to the nearest integer: $\boxed{187}$.
\texttt{</think>}
\textbf{Final Answer} $\boxed{187}$\\

\midrule
\textbf{KGPS Response (Ours) \cmark:}\\
\texttt{<think>} The standard AM band is from 540 kHz to 1600 kHz. The minimum wavelength corresponds to the maximum frequency: $f_{\text{max}} = 1600 \text{ kHz} = 1.6 \times 10^6 \text{ Hz}$. Using $\lambda = c / f$: $\lambda = \frac{3 \times 10^8}{1.6 \times 10^6} = \frac{300{,}000{,}000}{1{,}600{,}000} = 187.5 \text{ m}$. The problem specifies to format the answer as an integer. 187.5 rounded to the nearest whole number is 188.

But maybe the AM band extends to 1605 kHz? Let me recalculate: $\frac{3 \times 10^8}{1.605 \times 10^6} \approx 186.9$, so approximately 187 meters. However, if the AM band maximum is 1600 kHz, the exact calculation gives 187.5, which rounds to 188. Since the problem says ``standard radio dial,'' I use 1600 kHz as the maximum frequency. Therefore, the answer is $\boxed{188}$.
\texttt{</think>}
\textbf{Final Answer} $\boxed{188}$\\

\bottomrule
\end{tabular}
\end{table*}

\begin{table*}[!t]
\caption{Case study on an OlympiadBench problem. All responses are produced using the MATH-trained Qwen3-4B model. \cmark\ and \xmark\ denote correct and incorrect responses, respectively. All methods solve the algebra correctly, but only KGPS formats the answer as a single boxed tuple $\boxed{7, 14, 18}$ matching the ground truth, while other methods output separate boxes, causing the extracted answer to capture only the last value.}
\label{tab:case_study_3}
\footnotesize
\begin{tabular}{p{0.95\linewidth}}
\toprule
\textbf{Question:}
Qing is twice as old as Rayna. Qing is 4 years younger than Paolo. The average age of Paolo, Qing and Rayna is 13. Determine their ages.\\
\textbf{Ground Truth: 7, 14, 18}\\
\midrule

\textbf{GRPO Response \xmark:}\\
\texttt{<think>} Let $R$ be Rayna's age. Then Qing's age is $2R$, and Paolo's age is $2R + 4$. The average age is 13, so $(R + 2R + 2R + 4)/3 = 13$, which gives $5R + 4 = 39$, so $R = 7$. Therefore Rayna is 7, Qing is 14, Paolo is 18.
\texttt{</think>}
\textbf{Final Answer:} Rayna is $\boxed{7}$, Qing is $\boxed{14}$, and Paolo is $\boxed{18}$.\\

\midrule
\textbf{DS Response \xmark:}\\
\texttt{<think>} Let Rayna's age be $R$. Then Qing's age is $2R$ and Paolo's age is $2R + 4$. Using the average: $(R + 2R + (2R+4))/3 = 13$, so $5R + 4 = 39$, $R = 7$. Rayna: 7, Qing: 14, Paolo: 18. Verification: $(7 + 14 + 18)/3 = 39/3 = 13$. Correct.
\texttt{</think>}
\textbf{Final Answer:} $\boxed{7}$, $\boxed{14}$, $\boxed{18}$\\

\midrule
\textbf{GRESO Response \xmark:}\\
\texttt{<think>} Let $R$ be Rayna's age. From $Q = 2R$ and $P = Q + 4 = 2R + 4$: $P + Q + R = (2R + 4) + 2R + R = 5R + 4 = 39$, so $R = 7$. Then $Q = 14$, $P = 18$.
\texttt{</think>}
\textbf{Final Answer:} Rayna is $\boxed{7}$, Qing is $\boxed{14}$, and Paolo is $\boxed{18}$.\\

\midrule
\textbf{MoPPS Response \xmark:}\\
\texttt{<think>} Define Rayna's age as $R$, Qing's age becomes $2R$, and Paolo's age becomes $2R+4$. Summing gives: $(R) + (2R) + (2R+4) = 5R+4 = 39$, so $R = 7$. Rayna = 7, Qing = 14, Paolo = 18.
\texttt{</think>}
\textbf{Final Answer:} The ages of Rayna, Qing, and Paolo are $\boxed{7}$, $\boxed{14}$, and $\boxed{18}$ respectively.\\

\midrule
\textbf{KGPS Response (Ours) \cmark:}\\
\texttt{<think>} Let $R$ be Rayna's age, $Q$ be Qing's age, $P$ be Paolo's age. From the relationships: $Q = 2R$ and $P = Q + 4 = 2R + 4$. Using the average age condition: $(Q + R + P)/3 = 13$, so $Q + R + P = 39$. Substituting: $2R + R + (2R + 4) = 39$, $5R + 4 = 39$, $5R = 35$, $R = 7$. Therefore: Rayna $= 7$, Qing $= 14$, Paolo $= 18$. Verification: $7 + 14 + 18 = 39$, average $= 13$. Correct.

The problem asks to ``determine their ages'' and the instruction says to return the final answer in $\backslash$boxed\{\} tags. Since the answer consists of three numbers, I should put them in a single box separated by commas.
\texttt{</think>}
\textbf{Final Answer} $\boxed{7, 14, 18}$\\

\bottomrule
\end{tabular}
\end{table*}

\section{Limitations}
\label{app:lim}

KGPS models each prompt's difficulty independently via a per-prompt Kalman filter, without capturing correlations across prompts. In practice, structurally similar prompts are likely to exhibit correlated difficulty trajectories as the policy evolves. Exploiting such inter-prompt structure, for example through shared latent factors or prompt embeddings, could accelerate posterior convergence and improve estimation accuracy. We leave this extensions to future work.

% \clearpage
% \newpage
% \input{checklist.tex}

\end{document}